\pdfoutput=1

\documentclass[11pt]{article}
\PassOptionsToPackage{table}{xcolor}

\usepackage[final]{acl}
\usepackage{times}
\usepackage{latexsym}
\usepackage{tabularx}
\usepackage{pgfplotstable}
\usepackage{booktabs} 
\usepackage{multirow} 
\usepackage{array}
\usepackage{graphicx}
\usepackage{algorithmic}
\usepackage{algorithm}
\usepackage{makecell} 
\usepackage{amsmath}
\usepackage[english]{babel}
\usepackage{amsthm}

\usepackage{xcolor}
\usepackage{booktabs,multirow}
\usepackage{makecell}
\usepackage{tcolorbox}
\usepackage{enumitem}
\usepackage{amsmath}
\usepackage{titlesec}

\DeclareMathOperator*{\argmax}{arg\,max}

\usepackage{times}
\usepackage{enumitem}

\usepackage[T1]{fontenc}
\usepackage{pifont}
\usepackage[utf8]{inputenc}

\usepackage{microtype}

\usepackage{inconsolata}

\usepackage{graphicx}

\title{Derailer-Rerailer: Adaptive Verification for Efficient and Reliable Language Model Reasoning}

\author{
  Guangya Wan$^{1}$\thanks{These authors contributed equally to this work.},
  Yuqi Wu$^{2}$\footnotemark[1],
  Hao Wang$^{3}$,
  Shengming Zhao$^{2}$,
  Jie Chen$^{2}$\thanks{Corresponding author.},
  Sheng Li$^{1}$\footnotemark[2] \\
  $^1$School of Data Science, University of Virginia \\
  $^2$Department of Electrical and Computer Engineering, University of Alberta \\
  $^3$Beaconfire Solution Inc. \\
  \texttt{\{wxr9et,shengli\}@virginia.edu}, \texttt{\{yuqi14,jc65,shengmi1\}@ualberta.ca}, \\
  \texttt{haowang229@gmail.com} \\
  \footnotetext[2]{Corresponding author.}
}

\begin{document}
\maketitle
\begin{abstract}

Large Language Models (LLMs) have shown impressive reasoning capabilities, yet existing prompting methods face a critical trade-off: simple approaches often struggle with complex tasks and reasoning stability, while more sophisticated methods require multiple inferences and substantial computational resources, limiting their practical deployment. To address this challenge, we propose \textbf{Derailer-Rerailer}, a novel framework that adaptively balances reasoning accuracy and computational efficiency. At its core, our framework employs a lightweight \textbf{Derailer} mechanism to assess reasoning stability and selectively triggers an advanced \textbf{Rerailer} verification process only when necessary, thereby optimizing computational resource usage. Extensive evaluation across both open and closed-source models on more than 20 categories of mathematical, symbolic, and commonsense reasoning tasks demonstrates our framework's effectiveness: \textbf{Derailer-Rerailer} achieves significant accuracy improvements (8-11\% across various reasoning tasks) while maintaining 2-3 times better efficiency than existing verification methods, with particularly strong performance in mathematical and symbolic reasoning, offering a practical solution for enhancing LLM reasoning reliability while significantly reducing computational overhead~\footnote{Code available at \url{https://github.com/wan19990901/CoT_rerailer}}.

\end{abstract}

\section{Introduction}

\begin{figure}[!t]
\begin{center}
\includegraphics[width=1.0\linewidth]{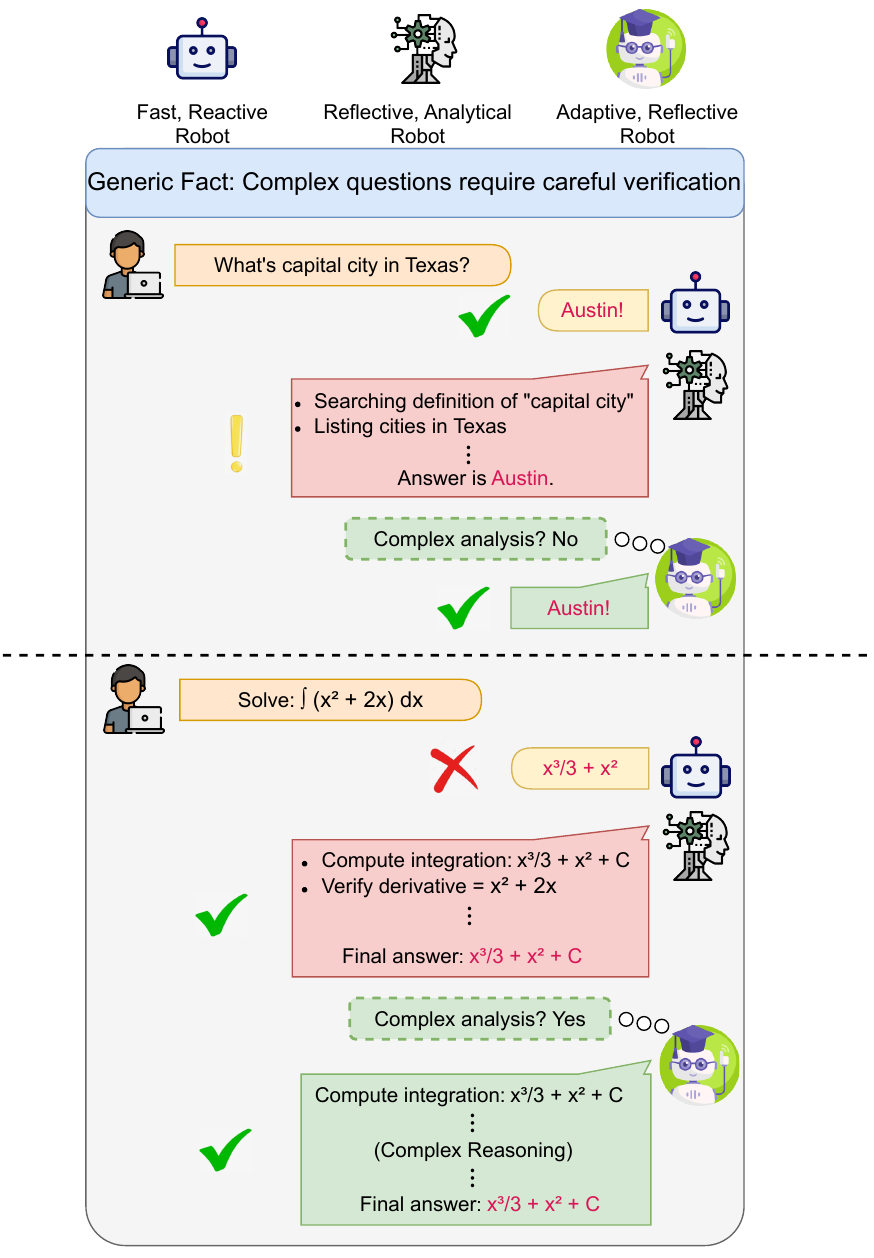}

\end{center}
\vspace{-5mm}
\caption{While simple questions can be answered quickly with minimal reasoning, complex tasks require deeper analysis and verification to ensure accuracy. Adaptive approaches balance efficiency and reasoning depth based on the problem's complexity.}
\label{fig: rerailer_overview}
\end{figure}

\begin{figure*}[!t]
\centering
\includegraphics[width=0.92\linewidth,height=0.82\textheight,keepaspectratio]{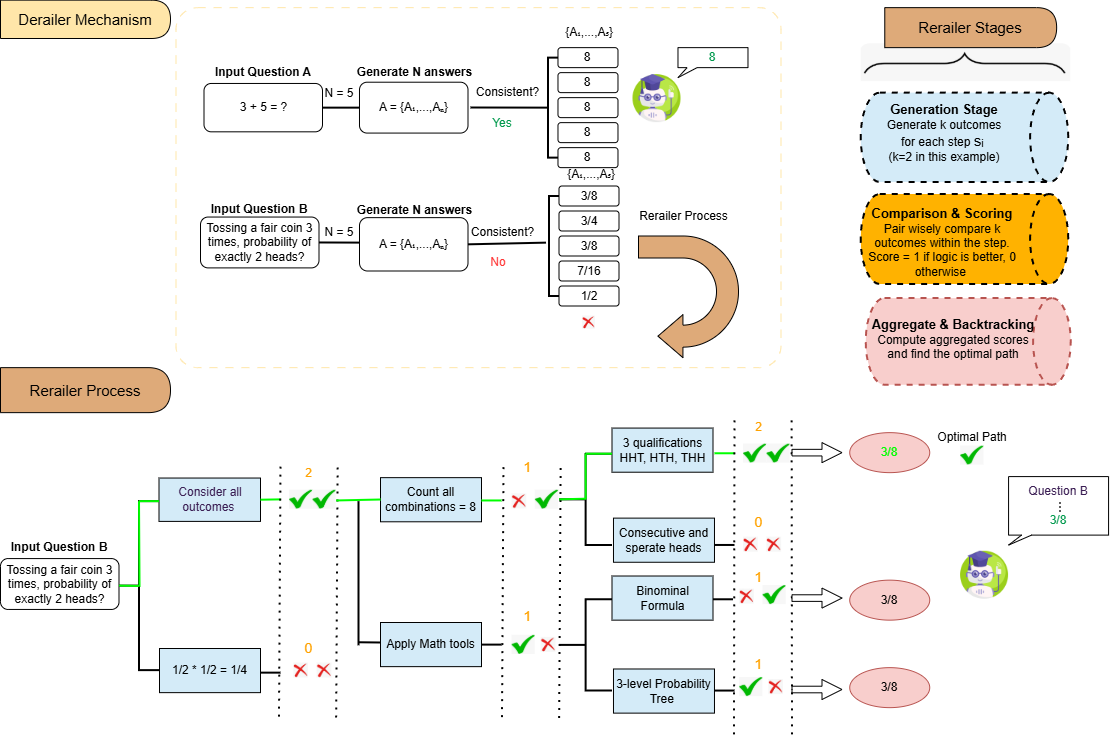}
\vspace{-4mm}
\caption{Overview of the Derailer-Rerailer framework: The Derailer filters questions based on reasoning stability, while the Rerailer optimizes reasoning paths for inconsistent cases. Arithmetic examples illustrate the workflow.}
\label{fig:rerailer_overview}
\end{figure*}

Large language models (LLMs) have been augmented with various prompting techniques to induce human-like complex reasoning capabilities \cite{wei2022emergent}. 
While early approaches like Chain-of-Thought aimed to emulate fast thinking for direct step-by-step reasoning \cite{wei2022chain}, they often struggled with complex multi-step reasoning and may produce unfaithful intermediate steps that propagate errors \cite{zhang2023sirens}.
Recent research has thus focused on enabling slower, more deliberate thinking in LLMs through various prompting strategies that allow exploration of multiple independent reasoning paths \cite{yao2024tree} or engagement in self-critique \cite{miao2024selfcheck}. This progression from fast to slow reasoning has demonstrated significant success in enhancing LLMs' reasoning capabilities, paralleling human cognitive processes where we transition from quick intuition to careful analysis when facing complex challenges \cite{kahneman2011thinking}.

These complex prompting methods, however, face a critical trade-off between accuracy and computational efficiency. Such approaches dramatically increase computational overhead. For example, in Self-Consistency \cite{wang2023selfconsistency}, a single problem requires 40 model calls and results in thousands of tokens \cite{li2024escape}. While this computational intensity proves necessary for some complex situations, a key limitation lies in their uniform application—indiscriminately deploying these expensive procedures across all queries regardless of whether the problem actually requires such intensive operations. This computational burden becomes particularly problematic in real-world agentic applications \cite{huang2024agentcodermultiagentbasedcodegeneration} that often involve various complex prompting methods, where inference time is as crucial \cite{snell2024scalingllmtesttimecompute} as accuracy (e.g. real-time clinical support agentic systems, which require both reliable reasoning and low latency \cite{umerenkov2023decipheringdiagnoseslargelanguage,wu2025wisemindrecontextualizingaiknowledgeguided}). These constraints highlight the need for an adaptive prompting method that can selectively apply computational resources where they are most needed.

To address these limitations, we propose a novel two-stage framework, \emph{Derailer-Rerailer} (Fig.~\ref{fig:rerailer_overview}), inspired by the relationship between LLMs' stability on answers and problem solvability.
Drawing an analogy from train operations—where derailment assessment precedes strategic rerailment—our framework combines two complementary mechanisms: (1) The \textbf{Derailer} performs full consistency checks through multiple independent answers, efficiently identifying which queries require intervention; and (2) the \textbf{Rerailer} applies targeted correction techniques only to cases where inconsistencies are detected. This selective approach ensures that expensive verification methods are deployed only when necessary, simultaneously improving accuracy and preserving computational efficiency.

\begin{table*}[h!]
\centering
\scriptsize
\caption{Comparison of prompting strategies that leverage both fast,reactive and slow, reflective reasoning modes in LLMs. 
Immediate prompting provides low QA improvement with low token usage and inference time. Whereas Iterative prompting yields higher QA improvement, but at the cost of higher token usage and inference time. 
Our goal in this paper is to combine the best of both.
}
\label{tab:prompts_compare}
\setlength{\tabcolsep}{8pt}
\renewcommand{\arraystretch}{1.2}
\begin{tabular}{llccc}
\toprule
\textbf{Category}  & \textbf{Example Methods} & \textbf{QA Perf.} & \textbf{Token Usage} & \textbf{Inference Time} \\
\midrule
\textbf{Immediate Prompting}      & \multirow{1}{4cm}{\emph{Chain of Thought, Least-to-Most}}                      & Low                     & Low                  & Low                     \\
\midrule
\multirow{2}{*}{\textbf{Iterative Prompting}}    & \multirow{2}{5cm}{\emph{Self-Consistency, Chain-of-Verification, \\     Tree of Thought, \textbf{Rerailer}}}    & \multirow{2}{*}{High}                    & \multirow{2}{*}{High}                 & \multirow{2}{*}{High}                    \\ 
\\
\midrule
\multirow{2}{*}{\textbf{Adaptive Iterative Prompting (Ours)}} & \multirow{2}{5cm}{\emph{Derailer + \{Self-Consistency, Chain-of-Verification, Tree of Thought, \textbf{Rerailer}\}}} & \multirow{2}{*}{\textbf{High}} & \multirow{2}{*}{\textbf{Medium}} & \multirow{2}{*}{\textbf{Medium}} \\ 
\\
\bottomrule
\end{tabular}
\end{table*}

We have conducted experiments across multiple LLMs on 7 reasoning benchmarks covering more than 20 categories to demonstrate the effectiveness of our framework. The result shows that our efficient framework outperforms various prompting methods up to 10\% in accuracy and by more than 50\% in efficiency metrics. The framework also shows robust performance across different reasoning types, with the most substantial improvements observed in mathematical and symbolic reasoning tasks. In summary, our contributions are as follows:

\begin{itemize}
    \item \textbf{Derailer Mechanism:} We introduce an efficient approach for detecting reasoning instability, enabling targeted application of complex prompting techniques.
    
    \item \textbf{Rerailer Verification:} We develop a novel prompting method that enhances reasoning stability while preserving accuracy and efficiency.
    
    \item \textbf{Practical Insights:} We provide comprehensive empirical evidence on the efficiency-accuracy trade-offs in various LLM prompting strategies, offering guidelines for deploying LLMs in resource-constrained environments.
\end{itemize}



    
    
    
    
    
    

\section{Motivations and Preliminary Findings}
\label{sec:methodology}


\begin{figure*}[!t]
\begin{center}
\label{fig:preliminary}
\includegraphics[width=1\linewidth]{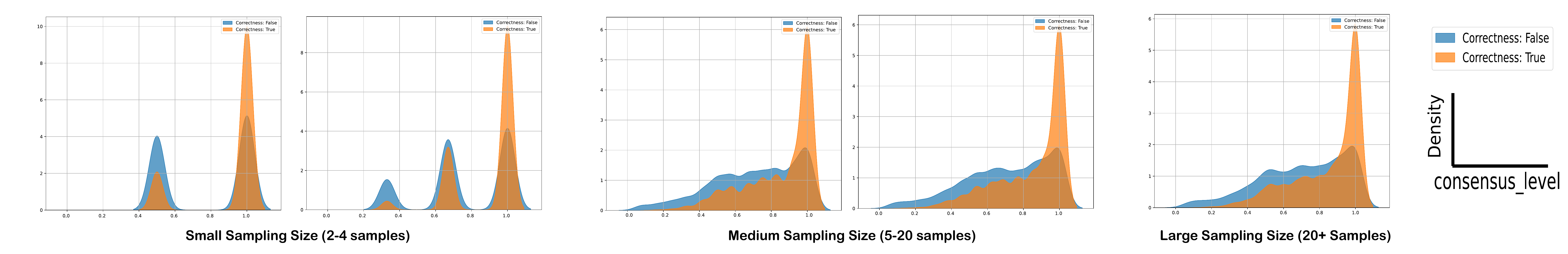}

\end{center}
\vspace{-5mm}
\caption{Impact of Sampling Size on Model Performance Consistency. The distributions show consensus levels for \emph{Solvable} questions (orange) and \emph{Insolvable} questions (blue). Distinct patterns on high consensus peaks for \emph{Consistently Solvable} questions and a notable peak for \emph{Consistently Insolvable} questions were observed, indicating that models maintain consistency both when questions are within and fundamentally beyond their capabilities. Besides, as sampling size increases (medium: 5-20, large: 20+), these patterns stabilize, suggesting that minimal sampling is sufficient to assess whether a question falls into our three-way classification.}
\label{fig: preliminary}
\end{figure*}

\subsection{Emulating Human Cognitive Systems in Language Model Prompting}

Human cognition, as described by Kahneman's dual-process theory \cite{gronchi2018dual}, operates through two distinct systems: System 1, which provides quick, intuitive responses, and System 2, which enables slower, more deliberate reasoning. While System 1 excels at routine tasks through its efficiency, System 2's complex approach proves invaluable for problems requiring careful analysis. This cognitive framework provides a valuable lens for understanding and improving language model reasoning, where different prompting techniques can be viewed as analogs to these two systems. In this paper, we will define the categories of prompting, aiming to balance the benefits of both fast and deliberate reasoning.

\noindent\textbf{Immediate Prompting:}
This represents the simplest form of interaction with a language model - a single-pass generation process. While it can still generate reasoning steps, the key characteristic is that it produces the answer in one forward pass:
\[
y = \mathcal{M}(x, p)
\]
where $p$ encapsulates the prompting strategy (such as Chain of Thought \cite{wei2022chain}, Least to Most \cite{zhou2023leasttomost}). Even when the output contains intermediate reasoning steps $s_1,...,s_n$, they flow in a single, uninterrupted sequence:
\[
{s_1,...,s_n,y} = \mathcal{M}(x, p)
\]
\noindent\textbf{Iterative Prompting:}
Another category is iterative prompting which embraces a more deliberate approach. It either generates \textit{parallel} independent samples (like Self-Consistency \cite{wang2023selfconsistency} or Best of N sampling \cite{openai2022goodhart}) and use some sort of aggregation \cite{wan2024dynamic} or selection \cite{gu2025surveyllmasajudge} mechanism to obtain the final answer, or \textit{sequentially} evaluates and refines the reasoning path based on some evaluators (like Chain of Verification \cite{dhuliawala2023chainofverification} or Self-Refine \cite{madaan2023selfrefine}) Fomrally, this can be defined as:
\[
y_i = \mathcal{M}(x,p_i) \quad \text{for} \quad i = 1,...,k
\]
\[
y^* = V({y_1,...,y_k}, x)
\]
where $V$ represents reward functions that leverage additional model calls to refine the initial solutions or explore other indepedent solution paths. With a better reasoning performance, the computational cost of iterative prompting increases significantly scaled by both the number of attempts $k$ and the cost of verification $V$.

The fact that human cognition does not necessitate deliberate processing for all decisions suggests a parallel for language models: computationally intensive iterative prompting may not always be advantageous.  This motivates the central question: \textbf{Under what conditions is the investment in additional computation for deliberate reasoning (Iterated Prompting) warranted?}  

Our proposed answer lies in analyzing the stability of the model's reasoning process.  We argue that the variability in LLM performance—ranging from high confidence to complete uncertainty, and including instances of knowledge without consistent application—suggests three distinct categories: \noindent\textbf{Stable Success}: When the model demonstrates consistent success with parallel sampling, usually in some single question like "what's the result of 1 + 2", engaging in additional refinement thus becomes computationally inefficient. \noindent\textbf{Stable Failure}: When the model consistently fails regardless of the approach, indicating a fundamental knowledge or reasoning gap that additional computation cannot bridge. \noindent\textbf{Unstable Reasoning}: When the model shows potential but lacks consistency, producing a mix of successful and failed attempts. These cases represent ideal candidates for a system 2 style of thinking, where multiple attempts can help stabilize the reasoning process.

\subsection{Empirical Validation}
To empirically validate the relationship between answer consistency and reasoning stability, we conducted a controlled study across 3 LLMs (Claude-3.5-sonnet, Llama3.1-70B, GPT-4o-mini) using datasets from various reasoning benchmarks including tasks where models typically exhibit stable reasoning (GSM8K, CommonsenseQA, etc.) and those that naturally lead to unstable outputs (such as TruthfulQA, GPQA-Diamond, etc.).

As shown in Fig.~\ref{fig: preliminary}, this analysis reveals two key patterns. First, regarding sampling efficiency, our experiments show that small samples already provide clear distinctions in consensus levels (the proportion of answers belonging to the majority class), with these patterns stabilizing at medium sampling sizes and larger samples offering diminishing returns. This aligns with recent findings \citep{wan2024dynamic} on the effectiveness of reduced sampling in self-consistency approaches, which suggests a small set of samples determines the solvability of question. Second, we observe distinct peaks in both distributions: while questions with correct reasoning (orange) show high stability as expected, we also find a notable peak for cases of \emph{incorrect reasoning} (blue) where questions consistently exceed the model's capabilities. This suggests that sampling just a few examples can effectively identify a question's stability category, thus motivating the design of an adaptive prompting method that adjusts computational investment based on reasoning stability, as suggested in Table \ref{tab:prompts_compare}.

\section{Methodology}
\subsection{Derailer: A Mechanism to Filter Stabilized Reasoning}

Building upon the above motivation and analysis, we propose Derailer, a selective gatekeeper that identifies and filters out cases with stable reasoning (whether consistently correct or incorrect) where expensive iterative prompting would provide little benefit. As shown in Fig.~\ref{fig: rerailer_overview}, Derailer implements this filtering through a lightweight consistency check with n samples. Only when samples yield inconsistent answers, indicating unstable reasoning patterns, does the question proceed to iterative prompting for deeper analysis. This filtering mechanism naturally aligns with our stability-based classification: cases with stable reasoning (both correct and incorrect) are filtered out to avoid unnecessary computation, while cases exhibiting reasoning instability receive additional verification and refinement.

The effectiveness of Derailer relies on two key conditions: First, the proportion of cases with stable reasoning should be substantial - a condition satisfied by modern LLMs which tend to be either consistently right or consistently wrong on many tasks as demonstrated in our preliminary studies and other work \cite{wan2024dynamic}. Second, the stability check should be computationally efficient by keeping the number of samples $n$ low. By meeting these conditions, Derailer optimizes the trade-off between computational cost and answer quality, applying more intensive reasoning procedures only when the model's unstable reasoning patterns suggest potential for improvement through iteration.

\subsection{Rerailer: Stabilizing Inconsistent Reasoning}

\paragraph{Motivation: }
While Derailer serves as a sampling-based diagnostic mechanism that can precede any iterative refinement algorithm, existing approaches aren't specifically designed to handle cases of unstable reasoning. Through analysis of examples like Figure \ref{postive_sample_physics}, we observe that models often make intermediate mistakes due to incorrect path selection and unstable execution of otherwise viable strategies. This observation motivates our Rerailer mechanism, which stabilizes inconsistent reasoning patterns through targeted pairwise comparisons—particularly beneficial for questions where LLMs demonstrate potential but lack consistency (as evidenced by the left-tailed distribution in Figure \ref{fig: preliminary}).

The core insight behind Rerailer is that unstable performance typically stems from error accumulation in multi-step reasoning, formalized as 
\begin{equation}
P(y, S|x) = P(s_1|x) \prod_{i=2}^{n} P(s_i|s_{1:i-1}, x) P(y|S, x)
\end{equation}

where errors at any intermediate step $s_i$ can propagate through conditional dependencies $P(s_i|s_{1:i-1},x)$, destabilizing the entire process even when the model possesses the fundamental capability. While approaches like Tree of Thoughts (ToT) similarly explore multiple decoding paths, Rerailer offers two key advantages: (1) the evaluation for each state employs pairwise comparisons between independent solutions rather than direct value assignment through few-shot learning, proving more reliable as models generally perform better at comparative judgments than absolute scoring; and (2) we employ a new search mechanism that dynamically explores branches only when comparisons yield ties, rather than maintaining a full tree exploration—enabling focused intervention precisely where reasoning instability occurs while preserving both accuracy and computational efficiency.

\paragraph{Framework and Implementation:}
As demonstrated on the bottom half of Fig \ref{fig: rerailer_overview}, The Rerailer implements an efficient approach to stabilize multi-step reasoning through four key stages:
\begin{enumerate}[leftmargin=*]
\renewcommand{\itemsep}{0.5mm}

\item \textbf{Candidate Generation:} For each reasoning step $i$, generate two candidate solutions using an adaptive sampling strategy:
\[\{s_i^1, s_i^2\} \sim \mathcal{M}(\cdot|x, s_{1:i-1})\]
The language model $\mathcal{M}$ is conditioned on the problem instance $x$ and the previously generated reasoning steps $s_{1:i-1}$.

\item \textbf{Pairwise Comparison:} Leverage the LLM's inherent reasoning capabilities to evaluate the semantic and logical differences between candidate thoughts. We employ a bidirectional voting mechanism where each pair is compared twice, with a scoring function $f(s_i^1, s_i^2)$ defined as:
\[
f(s_i^1, s_i^2) =
\begin{cases}
(2,0) & \text{if both favor } s_i^1 \text{ over } s_i^2 \\
(1,1) & \text{if yield a tie decision} \\
(0,2) & \text{if both favor } s_i^2 \text{ over } s_i^1
\end{cases}
\]
Each solution pair is evaluated independently in both directions to mitigate potential ordering bias and ensure more robust outcomes.

\item \textbf{Adaptive Strategy Selection:} Implement an adaptive decision mechanism based on the comparison outcomes:
\[\sigma(s_i) = \begin{cases}
\text{greedy} & f(s_i^1, s_i^2) = (2,0), (0,2) \\
\text{explore} & f(s_i^1, s_i^2) = (1,1)
\end{cases}\]
A greedy strategy is employed when both comparisons prefer one candidate, while an exploratory strategy is triggered when the comparisons yield a split decision, indicating reasoning instability that warrants exploration of both paths.

\textit{Greedy Extension:} If $\sigma(s_i) = \text{greedy}$, select the preferred candidate based on the pairwise comparison scores:
\[
s_i^* =
\begin{cases}
s_i^1 & \text{if } f(s_i^1, s_i^2) = 2 \\
s_i^2 & \text{if } f(s_i^1, s_i^2) = 0
\end{cases}
\]
Append $s_i^*$ to the current path for subsequent reasoning.

\textit{Exploratory Extension:} If $\sigma(s_i) = \text{exploratory}$, the tied votes ($f(s_i^1, s_i^2) = (1,1) $) indicate reasoning instability. Extend the current path in \textbf{both} directions, creating parallel paths to explore alternative reasoning routes. Both extensions are retained for subsequent reasoning.

The algorithm then returns to \textbf{(1) Candidate Generation} for the next reasoning step ($i+1$), using the extended path(s) as the new $s_{1:i}$. This iterative process continues until a complete solution is reached.

\item \textbf{Final Path (Answer) Selection:} After the iterative reasoning process concludes, evaluate all generated paths.  Calculate a score for each path $S = \{s_1, s_2, ..., s_n\}$ by summing the scores of the individual steps:
\[\text{Score}(S) = \sum_{i=1}^{n} f(s_i, \cdot)\]
Select the path with the highest score as the final reasoning chain $S^*$:
\[S^* = \argmax_S \text{Score}(S)\]

This selects the answer $y$ which is expected to be the last step of $S^*$.

\end{enumerate}

This structured process systematically improves reasoning stability for complex cases. By focusing on stabilizing intermediate steps, the Rerailer addresses the core challenge of \emph{unstable reasoning} questions left unsolved by the Derailer.

\begin{table*}[t]
\centering
\caption{Comparison of reasoning methods across multiple models and datasets. 
All accuracy metrics are expressed in percentages. 
\textbf{Math Acc} measures performance on mathematical reasoning problems. 
\textbf{Symbolic Acc} measures accuracy on symbolic reasoning tasks. 
\textbf{Commonsense Acc} reflects performance on common-sense reasoning tasks. 
\textbf{Overall Acc} is a combined measure across these categories.
\textbf{Tokens (K)} indicates the total number of tokens used (in thousands). 
\textbf{Acc/ K Token Gain (AGKT)} shows how much the accuracy improves per 1K tokens over that model's zero-shot CoT baseline. 
Values in parentheses next to the Overall Acc indicate the relative improvement compared to the zero-shot CoT baseline for each model.}
\label{tab:model_comparison_expanded}
\scriptsize
\renewcommand{\arraystretch}{1.2}
\begin{tabular}{@{}l@{\hspace{0.45em}}l@{\hspace{0.45em}}c@{\hspace{0.45em}}c@{\hspace{0.45em}}c@{\hspace{0.45em}}c@{\hspace{0.45em}}c@{\hspace{0.45em}}c@{}}
\toprule
\textbf{Model} & \textbf{Method} & \makecell{\textbf{Math Acc} (\%)} & \makecell{\textbf{Symbolic}\\\textbf{Acc} (\%)} & \makecell{\textbf{Commonsense}\\\textbf{Acc} (\%)} & \makecell{\textbf{Overall Acc} (\%)} & \makecell{\textbf{Tokens (K)}} & \makecell{\textbf{Acc Gain per}\\\textbf{K Tokens (\%)}} \\
\midrule
\multirow{11}{*}{Claude-3.5-Sonnet} 
 & Zero-shot CoT & 68.3 & 77.6 & 72.2 & 72.7 & 0.108 & - \\
 & Least-to-Most CoT & 70.4 & 79.3 & 74.1 & 74.6 \textcolor{green!50!black}{(+1.9)} & 0.372 & 5.11 \\
 & Five-shot CoT & 72.9 & 81.4 & 74.8 & 76.4 \textcolor{green!50!black}{(+3.7)} & 0.518 & 7.14 \\
 & Self-Consistency(SC) & 77.8 & 86.5 & 75.2 & 79.8 \textcolor{green!50!black}{(+7.1)} & 1.732 & 4.10 \\
 & Derailer + SC & 77.6 & 86.3 & 75.1 & 79.7 \textcolor{green!50!black}{(+7.0)} & 0.762 & 9.19 \\
 & Chain of Ver (CoVe) & 78.2 & 86.8 & 75.4 & 80.1 \textcolor{green!50!black}{(+7.4)} & 1.778 & 4.16 \\
 & Derailer + CoVe & 77.9 & 86.6 & 75.3 & 79.9 \textcolor{green!50!black}{(+7.2)} & 0.793 & 9.08 \\
 & Tree of Thought (ToT) & 78.1 & 86.7 & 75.4 & 80.1 \textcolor{green!50!black}{(+7.4)} & 2.245 & 3.30 \\
 & Derailer + ToT & 77.8 & 86.5 & 75.2 & 79.8 \textcolor{green!50!black}{(+7.1)} & 0.845 & 8.40 \\
 & Rerailer & 78.0 & 86.7 & 75.3 & 80.0 \textcolor{green!50!black}{(+7.3)} & 1.458 & 5.01 \\
 & \cellcolor{lightgray!30}Derailer + Rerailer & \cellcolor{lightgray!30}\textbf{80.8} & \cellcolor{lightgray!30}\textbf{89.7} & \cellcolor{lightgray!30}\textbf{75.9} & \cellcolor{lightgray!30}\textbf{82.1} \textcolor{green!50!black}{(+9.4)} & \cellcolor{lightgray!30}0.592 & \cellcolor{lightgray!30}\textbf{15.88} \\
\midrule
\multirow{11}{*}{Llama-3.1 70B} 
 & Zero-shot CoT & 38.2 & 71.3 & 70.4 & 60.0 & 0.132 & - \\
 & Least-to-Most CoT & 39.6 & 72.4 & 72.3 & 61.4 \textcolor{green!50!black}{(+1.4)} & 0.448 & 3.13 \\
 & Five-shot CoT & 41.8 & 74.7 & 72.9 & 63.1 \textcolor{green!50!black}{(+3.1)} & 0.628 & 4.94 \\
 & Self-Consistency(SC) & 45.9 & 79.1 & 73.4 & 66.1 \textcolor{green!50!black}{(+6.1)} & 1.932 & 3.16 \\
 & Derailer + SC & 45.7 & 78.8 & 73.3 & 65.9 \textcolor{green!50!black}{(+5.9)} & 0.892 & 6.61 \\
 & Chain of Ver (CoVe) & 46.2 & 79.4 & 73.6 & 66.4 \textcolor{green!50!black}{(+6.4)} & 1.972 & 3.25 \\
 & Derailer + CoVe & 46.0 & 79.2 & 73.5 & 66.2 \textcolor{green!50!black}{(+6.2)} & 0.923 & 6.72 \\
 & Tree of Thought (ToT) & 46.1 & 79.3 & 73.5 & 66.3 \textcolor{green!50!black}{(+6.3)} & 2.458 & 2.56 \\
 & Derailer + ToT & 45.9 & 79.1 & 73.4 & 66.1 \textcolor{green!50!black}{(+6.1)} & 0.968 & 6.30 \\
 & Rerailer & 46.1 & 79.3 & 73.5 & 66.3 \textcolor{green!50!black}{(+6.3)} & 1.652 & 3.81 \\
 & \cellcolor{lightgray!30}Derailer + Rerailer & \cellcolor{lightgray!30}\textbf{48.4} & \cellcolor{lightgray!30}\textbf{81.8} & \cellcolor{lightgray!30}\textbf{74.2} & \cellcolor{lightgray!30}\textbf{68.1} \textcolor{green!50!black}{(+8.1)} & \cellcolor{lightgray!30}0.648 & \cellcolor{lightgray!30}\textbf{12.50} \\
\midrule
\multirow{11}{*}{GPT-4o-mini} 
 & Zero-shot CoT & 59.5 & 46.8 & 69.2 & 58.5 & 0.121 & - \\
 & Least-to-Most CoT & 62.4 & 49.1 & 71.8 & 61.1 \textcolor{green!50!black}{(+2.6)} & 0.389 & 6.68 \\
 & Five-shot CoT & 63.9 & 50.8 & 72.4 & 62.4 \textcolor{green!50!black}{(+3.9)} & 0.556 & 7.01 \\
 & Self-Consistency (SC) & 68.6 & 55.2 & 72.9 & 65.6 \textcolor{green!50!black}{(+7.1)} & 1.795 & 3.96 \\
 & Derailer + SC & 68.3 & 54.9 & 72.8 & 65.3 \textcolor{green!50!black}{(+6.8)} & 0.823 & 8.26 \\
 & Chain of Ver (CoVe) & 69.3 & 55.5 & 73.2 & 66.0 \textcolor{green!50!black}{(+7.5)} & 1.842 & 4.07 \\
 & Derailer + CoVe & 68.9 & 55.3 & 73.1 & 65.8 \textcolor{green!50!black}{(+7.3)} & 0.864 & 8.45 \\
 & Tree of Thought (ToT) & 69.2 & 55.4 & 73.1 & 65.9 \textcolor{green!50!black}{(+7.4)} & 2.325 & 3.18 \\
 & Derailer + ToT & 68.8 & 55.2 & 73.0 & 65.7 \textcolor{green!50!black}{(+7.2)} & 0.912 & 7.89 \\
 & Rerailer & 69.0 & 55.4 & 73.1 & 65.8 \textcolor{green!50!black}{(+7.3)} & 1.539 & 4.74 \\
 & \cellcolor{lightgray!30}Derailer + Rerailer & \cellcolor{lightgray!30}\textbf{72.0} & \cellcolor{lightgray!30}\textbf{58.5} & \cellcolor{lightgray!30}\textbf{73.8} & \cellcolor{lightgray!30}\textbf{68.1} \textcolor{green!50!black}{(+9.6)} & \cellcolor{lightgray!30}0.639 & \cellcolor{lightgray!30}\textbf{15.02} \\
\bottomrule
\end{tabular}
\end{table*}

\section{Experiments}

We evaluate our framework across three broad categories: Mathematical, Symbolic, and Commonsense Reasoning, using 27 data categories from a total of 7 standard benchmarks (e.g BigBenchHard, MATH, StrategyQA, etc.) to ensure comprehensive coverage of real-world reasoning challenges. Experiments are conducted using three open and closed source LLMs with fixed temperature of 0.5, comparing against several established baselines that represent different paradigms in multi-step reasoning: Least to Most, Chain-of-Thought and its variants represent immediate prompting approaches that generate reasoning paths in a single forward pass; We also explore (1) Self-Consistency (SC) explores multiple solution paths independently; (2) Chain-of-Verification (CoVe) focuses on sequentially improving a single reasoning path through step-wise validation; and (3) Tree-of-Thought (ToT) combines both parallel exploration and sequential refinement through a tree-structured search as three baselines in iterative prompting. These selected methods represent fundamental methods that imply applicability to many advanced techniques building upon these methods \cite{liu-etal-2024-era,yao-etal-2024-got}

To evaluate the performance of different prompting methods, we evaluate (1) \textbf{\textit{effectiveness}}: through accuracy across different reasoning types, and (2) \textbf{\textit{efficiency}}: through total token consumption (sum of input and output tokens). 
To better show and evaluate the trade-off, we additionally introduce a novel metric, \textit{\textbf{Accuracy Gain per K Token (AGKT)}}, which quantifies the efficiency of prompting methods by measuring accuracy improvement per 1K tokens relative to the zero-shot baseline. Unlike time and number of API calls, this metric provides a Model and hardware independent assessment of computational efficiency, offering more reliable comparisons than direct runtime measurements that can vary across different settings.

\subsection{Main Results}
We analyze the experimental results from Table~\ref{tab:model_comparison_expanded} along several dimensions:

\noindent\textbf{Performance Gains over Various Prompting.} 
Our Derailer-Rerailer framework consistently outperforms all baseline methods across all evaluated models, demonstrating superior performance in both accuracy and computational efficiency. This synergistic combination leverages the strengths of both components: Derailer's ability to reduce computational overhead and Rerailer's effectiveness in error correction, resulting in a robust and efficient reasoning system.

\noindent\textbf{Excelling on Math and Symbolic Reasoning Tasks.} 
Analysis reveals distinct patterns of improvement across different reasoning types. The framework shows particularly strong performance in mathematical and symbolic reasoning tasks, while improvements in commonsense reasoning are more moderate. This disparity aligns with empirical expectations, as structured reasoning tasks typically benefit more from explicit verification processes. The smaller gains in commonsense reasoning can be attributed to the inherent ambiguity in these tasks, where determining the validity of intermediate steps requires more nuanced evaluation.

\noindent\textbf{Efficiency Gain with Iterative Prompting.} 
While Iterative prompting methods like Self-Consistency demonstrate accuracy improvements over immediate prompting, they come at a substantial computational cost. Our Derailer component, when integrated with these complex iterative prompting methods, maintains comparable accuracy while significantly reducing token consumption. This demonstrates that strategic, selective verification can be more resource-efficient than comprehensive checking approaches without compromising performance.


\begin{figure}[t]
\begin{center}
\includegraphics[width = \linewidth]{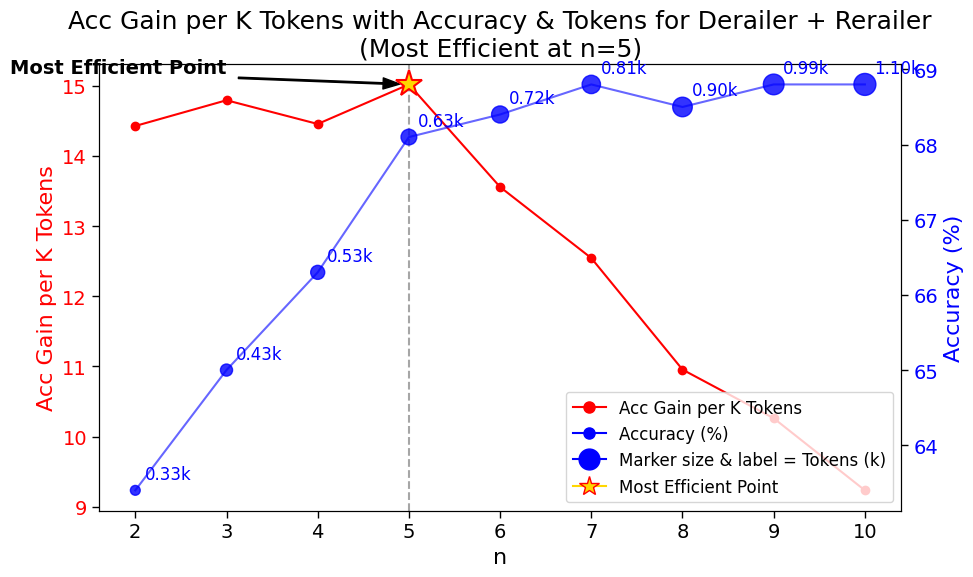}
\end{center}
\vspace{-4mm}
\caption{Analysis of sample size ($n$) effects in the Derailer-Rerailer framework. The blue line (right y-axis) shows accuracy improvement with increasing samples, with marker size indicating token consumption. The red line (left y-axis) shows efficiency measured by accuracy gain per thousand tokens. The star marks the optimal efficiency point at $n=5$. Results are averaged across all models on math tasks}
\label{fig:sample_derailer}
\end{figure}

\subsection{Abalation Study and Analysis}

\paragraph{Sample Size - Derailer Analysis: } We conduct a detailed analysis of how the number of samples $n$ in the Derailer affects the overall framework performance. Fig.~\ref{fig:sample_derailer} shows the relationship between three key metrics: accuracy, token consumption, and accuracy gain per thousand tokens (AGTK). While accuracy continues to improve with more samples, increasing from 64.3\% at $n$ = 2 to 69\% at $n$ = 10, the token consumption grows linearly with $n$. The efficiency metric AGTK reveals a clear optimal point at $n$ = 5, where the framework achieves the best balance between accuracy improvement and computational cost. Beyond this point, the marginal accuracy gains (less than 1\% per additional sample) do not justify the increased token consumption, as evidenced by the sharp decline in AGTK after $n$ = 5.

We did additional analysis in Appendix \ref{APP:derailer} to reveal that the pass rates (percentage of consistency) of Derailer remain stable at around 50\% across different n, suggesting that approximately half of the questions are identified as either \emph{Consistently Solvable} or \emph{Intrinsically Insolvable}. This stability aligns with our preliminary findings about the relationship between answer consistency and question solvability, providing additional evidence that a small number of samples is sufficient for reliable classification.

\begin{table}[t]
\centering
\caption{Comparison of Different Preference Ranking Methods and Number of Samples Generated at Each Step $k$. Accuracy in \% and Tokens in K. Results are averaged crossed all models}
\label{tab:sample_expansion}
\scriptsize
\setlength{\tabcolsep}{1.8pt}
\begin{tabular}{l c c c c c c@{}}
\toprule
\multirow{3}{*}{\textbf{Task}} & \multicolumn{2}{c}{\textbf{Binary ($k=2$)}} & \multicolumn{2}{c}{\textbf{Ranking ($k=3$)}} & \multicolumn{2}{c}{\textbf{Pairwise ($k=3$)}} \\
\cmidrule(lr){2-3}\cmidrule(lr){4-5}\cmidrule(lr){6-7}
 & Acc (\%) & Token (K) & Acc (\%) & Token (K) & Acc (\%) & Token (K) \\
\midrule
Math & 67.2 & 1.59 & 65.8 & 2.12 & 67.3 & 2.88 \\
Symbolic & 75.2 & 1.52 & 73.5 & 2.08 & 75.1 & 2.72 \\
Commonsense & 71.2 & 1.40 & 70.8 & 1.85 & 71.2 & 2.25 \\
Avg & 70.8 & 1.47 & 70.2 & 2.02 & 70.9 & 2.61 \\
\bottomrule
\end{tabular}
\end{table}

\begin{figure*}[t]
\begin{center}
\includegraphics[width = \linewidth]{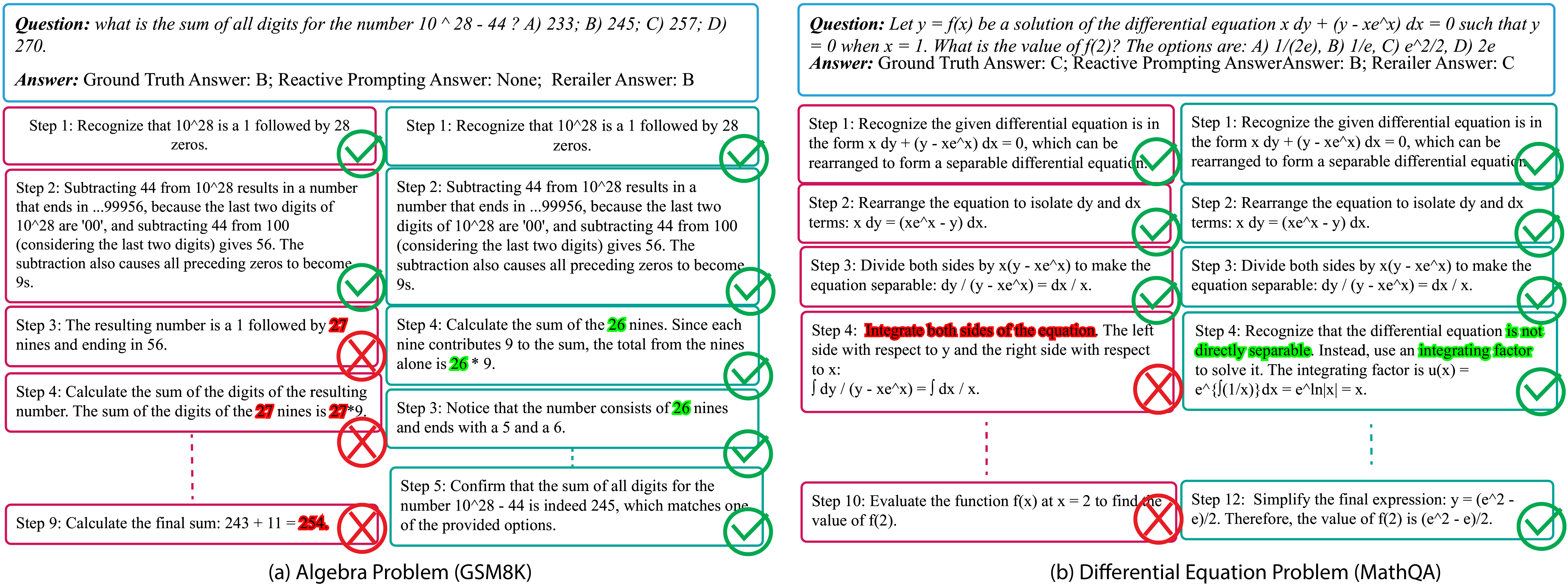}

\end{center}
\vspace{-4mm}
\caption{Derailer-Rerailer Solving Math Questions. The questions and answers, which were retrieved from the GSM8K and MathQA datasets, are exhibited in the blue box. The red boxes are steps generated via the baseline CoT method and the green boxes are the corrected RP from Rerailer. Mistakes are highlighted in red and corrections are highlighted in green. Additional examples across different problem types can be found in Appendix \ref{app:error_analysis}. }
\label{postive_sample_math}
\end{figure*}

\paragraph{Sample Size and Comparison Mechanism - Rerailer Analysis:}
We investigated two key aspects of the Rerailer framework: expanding the number of samples generated at each step $k$ and varying the comparison mechanism. First, we explored increasing the number of samples from $k=2$ to $k=3$, while maintaining the pairwise comparison approach. For a sample size $k$, the pairwise comparison requires $\binom{k}{2}$ comparisons, leading to $O(k^2)$ complexity in terms of model queries. This quadratic growth in computational cost is reflected in our empirical results: increasing from $k=2$ to $k=3$ maintains similar accuracy levels but incurs a 77\% increase in token consumption. We also examined an alternative ranking-based approach where the model directly orders $k=3$ samples simultaneously. While this approach achieves $O(k)$ complexity, it leads to decreased performance across all reasoning tasks. This superiority of pairwise comparisons, despite their higher computational cost, aligns with recent findings that language models excel at \emph{comparative judgments rather than holistic ranking tasks}, which is often biased by positional \cite{ wang2024directjudgementpreferenceoptimization, liu-etal-2024-lost}. The results further demonstrate the design choice of the Rerailer by leveraging the model's comparative reasoning strengths while keeping computational costs manageable.

\paragraph{Case Study:}  
 As suggested in Table \ref{tab:model_comparison_expanded}, Our Framework particularly excels in correcting flawed chains of thought in mathematical reasoning, addressing the key challenges of unstabilized reasoning patterns. Fig.\ref{postive_sample_math} demonstrates the Rerailer's effectiveness in identifying and rectifying errors in both basic and advanced math problems. For instance, in a counting problem, the Rerailer successfully stabilizes divergent reasoning paths by correcting computational mistakes that would have led to an incorrect final answer. Similarly, in a differential equation example, it identifies fundamental methodological errors and realigns the solution approach with correct mathematical principles. This correction mechanism directly addresses the common types of unstabilized reasoning: it prevents solution drift, resolves conflicting intermediate results, and remedies incomplete reasoning chains. The core capability of the Rerailer lies in stabilizing these patterns through systematic verification of intermediate steps, particularly benefiting tasks that can be \emph{decomposed into distinct, verifiable components}.
\section{Related Work}

\subsection{Prompting LLMs for Reasoning}

In-context Learning \cite{dong-etal-2024-survey}, including few-shot Chain-of-Thought prompting \cite{wei2022chain}, marked a breakthrough in eliciting reasoning capabilities from language models by encouraging them to articulate intermediate thought steps. Building upon this insight, researchers have developed various sophisticated methods to enhance reasoning performance. These approaches include different thinking style decompositions \cite{zhou2023leasttomost}, exploring multiple parallel reasoning paths \cite{wang2023selfconsistency, yao2024tree}, and sequential verification or refinement of generated rationales \cite{madaan2023selfrefine, dhuliawala2023chainofverification}, or combinations thereof \cite{yao2024tree, snell2024scalingllmtesttimecompute}. Other techniques, particularly those related to LLM agents \cite{Wang_2024}, rely on external knowledge or tool integration \cite{shinn2023reflexion, gao2023pal, yao2023react}, while additional training methods can also enhance reasoning capabilities \cite{trung-etal-2024-reft}. However, our paper focuses specifically on inference-time elicitation techniques for reasoning abilities already present in base models, which serve as the foundations for these more complex methods.

\subsection{Efficiency in Large Language Model Inference}

Despite significant advancements in reasoning capabilities through various prompting techniques, a critical challenge limiting the practical deployment of LLMs is their computational inefficiency \cite{wang2025harnessingreasoningeconomysurvey}. These models frequently exhibit unproductive behaviors, including generating redundant verifications after reaching correct answers or pursuing "fake thinking" patterns that simulate progress without meaningful problem-solving \cite{wu2025when, wang2025thoughts}, misallocating computational resources by applying uniform algorithms regardless of problem complexity or by inadequately distributing computation between simple and complex tasks \cite{wan2024dynamic,parashar2025inference, yang2025towards}. While existing solutions typically rely on sophisticated predictive strategies for budget allocation and algorithm selection \cite{han2025token, snell2024scaling}, we propose a simpler, more principled approach based on model's consistency for the input queries with (1) a Derailer mechanism that enhances efficiency by terminating unproductive reasoning paths, and (2) a Rerailer component that recovers accuracy in cases where the Derailer incorrectly intervenes.

\section{Conclusion}

In this work, we proposed \textbf{Derailer-Rerailer}, a novel two-stage framework that balances LLM reasoning accuracy and computational efficiency. Our framework demonstrates that selective verification can maintain the benefits of exhaustive reasoning while significantly reducing computational costs. Beyond the technical contribution, our findings reveal important practical implications: while increased computation generally improves accuracy, this relationship isn't linear, and many methods reach diminishing returns despite substantial resource investment. As LLMs continue to grow in both size and capability, we advocate for a more holistic evaluation approach that considers both accuracy and efficiency metrics for future research. This shift in evaluation criteria is particularly crucial for advancing research on LLM deployment in resource-constrained environments.

\section*{Limitations}

Our framework demonstrates promising results, but several important limitations warrant discussion:

\noindent\textbf{Preliminary Stability Analysis:} Our investigation of the relationship between the model's capability and answer consistency mainly motivates the design of our framework and remains high-level. A deeper theoretical understanding of how this relationship varies across reasoning tasks, model architectures, and knowledge domains could yield more nuanced strategies for applying iterative prompting adaptively.

\noindent\textbf{Sampling Parameter Selection:} Determining the optimal number of samples for the Derailer stage remains challenging, as this parameter significantly impacts both efficiency and effectiveness. Currently, this depends on model's pre-training and requires empirical tuning, which may not be practical in all deployment scenarios.

\noindent\textbf{Single Model Constraints:} The scope of this paper is to elicit the reasoning of a single model without external sources, extending the discussions of efficiency on understanding the recently developed reasoning models \cite{deepseekai2025deepseekr1incentivizingreasoningcapability} and exploring applications related to LLM's agents \cite{li2023camel} would be essential future directions.

\section*{Acknowledgments}
The work is supported in part by the National Science Foundation under Grants IIS-2316306 and CNS-2330215. The authors acknowledge Research Computing at The University of Virginia for providing computational resources and technical support that have contributed to the results reported within this publication.

\bibliography{custom}

\begin{thebibliography}{48}
\providecommand{\natexlab}[1]{#1}

\bibitem[{Amini et~al.(2019)Amini, Gabriel, Lin, Koncel-Kedziorski, Choi, and Hajishirzi}]{amini2019mathqa}
Aida Amini, Saadia Gabriel, Shanchuan Lin, Rik Koncel-Kedziorski, Yejin Choi, and Hannaneh Hajishirzi. 2019.
\newblock \href {https://doi.org/10.18653/v1/N19-1245} {Mathqa: Towards interpretable math word problem solving with operation-based formalisms}.
\newblock In \emph{Proceedings of the 2019 Conference of the North American Chapter of the Association for Computational Linguistics: Human Language Technologies, Volume 1 (Long and Short Papers)}, pages 2357--2367, Minneapolis, Minnesota. Association for Computational Linguistics.

\bibitem[{Anthropic(2024)}]{claude}
Anthropic. 2024.
\newblock The claude 3 model family: Opus, sonnet, haiku.

\bibitem[{Cobbe et~al.(2021)Cobbe, Kosaraju, Bavarian, Chen, Jun, Kaiser, Plappert, Tworek, Hilton, Nakano, Hesse, and Schulman}]{cobbe2021training}
Karl Cobbe, Vineet Kosaraju, Mohammad Bavarian, Mark Chen, Heewoo Jun, Lukasz Kaiser, Matthias Plappert, Jerry Tworek, Jacob Hilton, Reiichiro Nakano, Christopher Hesse, and John Schulman. 2021.
\newblock \href {https://arxiv.org/abs/2110.14168} {Training verifiers to solve math word problems}.
\newblock \emph{Preprint}, arXiv:2110.14168.

\bibitem[{DeepSeek-AI et~al.(2025)DeepSeek-AI, Guo, Yang, Zhang, Song, Zhang, Xu, Zhu, Ma, Wang, Bi, Zhang, Yu, Wu, Wu, Gou, Shao, Li, Gao, Liu, Xue, Wang, Wu, Feng, Lu, Zhao, Deng, Zhang, Ruan, Dai, Chen, Ji, Li, Lin, Dai, Luo, Hao, Chen, Li, Zhang, Bao, Xu, Wang, Ding, Xin, Gao, Qu, Li, Guo, Li, Wang, Chen, Yuan, Qiu, Li, Cai, Ni, Liang, Chen, Dong, Hu, Gao, Guan, Huang, Yu, Wang, Zhang, Zhao, Wang, Zhang, Xu, Xia, Zhang, Zhang, Tang, Li, Wang, Li, Tian, Huang, Zhang, Wang, Chen, Du, Ge, Zhang, Pan, Wang, Chen, Jin, Chen, Lu, Zhou, Chen, Ye, Wang, Yu, Zhou, Pan, Li, Zhou, Wu, Ye, Yun, Pei, Sun, Wang, Zeng, Zhao, Liu, Liang, Gao, Yu, Zhang, Xiao, An, Liu, Wang, Chen, Nie, Cheng, Liu, Xie, Liu, Yang, Li, Su, Lin, Li, Jin, Shen, Chen, Sun, Wang, Song, Zhou, Wang, Shan, Li, Wang, Wei, Zhang, Xu, Li, Zhao, Sun, Wang, Yu, Zhang, Shi, Xiong, He, Piao, Wang, Tan, Ma, Liu, Guo, Ou, Wang, Gong, Zou, He, Xiong, Luo, You, Liu, Zhou, Zhu, Xu, Huang, Li, Zheng, Zhu, Ma, Tang, Zha, Yan, Ren, Ren, Sha, Fu, Xu, Xie, Zhang,
  Hao, Ma, Yan, Wu, Gu, Zhu, Liu, Li, Xie, Song, Pan, Huang, Xu, Zhang, and Zhang}]{deepseekai2025deepseekr1incentivizingreasoningcapability}
DeepSeek-AI, Daya Guo, Dejian Yang, Haowei Zhang, Junxiao Song, Ruoyu Zhang, Runxin Xu, Qihao Zhu, Shirong Ma, Peiyi Wang, Xiao Bi, Xiaokang Zhang, Xingkai Yu, Yu~Wu, Z.~F. Wu, Zhibin Gou, Zhihong Shao, Zhuoshu Li, Ziyi Gao, Aixin Liu, Bing Xue, Bingxuan Wang, Bochao Wu, Bei Feng, Chengda Lu, Chenggang Zhao, Chengqi Deng, Chenyu Zhang, Chong Ruan, Damai Dai, Deli Chen, Dongjie Ji, Erhang Li, Fangyun Lin, Fucong Dai, Fuli Luo, Guangbo Hao, Guanting Chen, Guowei Li, H.~Zhang, Han Bao, Hanwei Xu, Haocheng Wang, Honghui Ding, Huajian Xin, Huazuo Gao, Hui Qu, Hui Li, Jianzhong Guo, Jiashi Li, Jiawei Wang, Jingchang Chen, Jingyang Yuan, Junjie Qiu, Junlong Li, J.~L. Cai, Jiaqi Ni, Jian Liang, Jin Chen, Kai Dong, Kai Hu, Kaige Gao, Kang Guan, Kexin Huang, Kuai Yu, Lean Wang, Lecong Zhang, Liang Zhao, Litong Wang, Liyue Zhang, Lei Xu, Leyi Xia, Mingchuan Zhang, Minghua Zhang, Minghui Tang, Meng Li, Miaojun Wang, Mingming Li, Ning Tian, Panpan Huang, Peng Zhang, Qiancheng Wang, Qinyu Chen, Qiushi Du, Ruiqi Ge, Ruisong
  Zhang, Ruizhe Pan, Runji Wang, R.~J. Chen, R.~L. Jin, Ruyi Chen, Shanghao Lu, Shangyan Zhou, Shanhuang Chen, Shengfeng Ye, Shiyu Wang, Shuiping Yu, Shunfeng Zhou, Shuting Pan, S.~S. Li, Shuang Zhou, Shaoqing Wu, Shengfeng Ye, Tao Yun, Tian Pei, Tianyu Sun, T.~Wang, Wangding Zeng, Wanjia Zhao, Wen Liu, Wenfeng Liang, Wenjun Gao, Wenqin Yu, Wentao Zhang, W.~L. Xiao, Wei An, Xiaodong Liu, Xiaohan Wang, Xiaokang Chen, Xiaotao Nie, Xin Cheng, Xin Liu, Xin Xie, Xingchao Liu, Xinyu Yang, Xinyuan Li, Xuecheng Su, Xuheng Lin, X.~Q. Li, Xiangyue Jin, Xiaojin Shen, Xiaosha Chen, Xiaowen Sun, Xiaoxiang Wang, Xinnan Song, Xinyi Zhou, Xianzu Wang, Xinxia Shan, Y.~K. Li, Y.~Q. Wang, Y.~X. Wei, Yang Zhang, Yanhong Xu, Yao Li, Yao Zhao, Yaofeng Sun, Yaohui Wang, Yi~Yu, Yichao Zhang, Yifan Shi, Yiliang Xiong, Ying He, Yishi Piao, Yisong Wang, Yixuan Tan, Yiyang Ma, Yiyuan Liu, Yongqiang Guo, Yuan Ou, Yuduan Wang, Yue Gong, Yuheng Zou, Yujia He, Yunfan Xiong, Yuxiang Luo, Yuxiang You, Yuxuan Liu, Yuyang Zhou, Y.~X. Zhu,
  Yanhong Xu, Yanping Huang, Yaohui Li, Yi~Zheng, Yuchen Zhu, Yunxian Ma, Ying Tang, Yukun Zha, Yuting Yan, Z.~Z. Ren, Zehui Ren, Zhangli Sha, Zhe Fu, Zhean Xu, Zhenda Xie, Zhengyan Zhang, Zhewen Hao, Zhicheng Ma, Zhigang Yan, Zhiyu Wu, Zihui Gu, Zijia Zhu, Zijun Liu, Zilin Li, Ziwei Xie, Ziyang Song, Zizheng Pan, Zhen Huang, Zhipeng Xu, Zhongyu Zhang, and Zhen Zhang. 2025.
\newblock \href {https://arxiv.org/abs/2501.12948} {Deepseek-r1: Incentivizing reasoning capability in llms via reinforcement learning}.
\newblock \emph{Preprint}, arXiv:2501.12948.

\bibitem[{Dhuliawala et~al.(2023)Dhuliawala, Komeili, Xu, Raileanu, Li, Celikyilmaz, and Weston}]{dhuliawala2023chainofverification}
Shehzaad Dhuliawala, Mojtaba Komeili, Jing Xu, Roberta Raileanu, Xian Li, Asli Celikyilmaz, and Jason Weston. 2023.
\newblock \href {https://arxiv.org/abs/2309.11495} {Chain-of-verification reduces hallucination in large language models}.
\newblock \emph{Preprint}, arXiv:2309.11495.

\bibitem[{Dong et~al.(2024)Dong, Li, Dai, Zheng, Ma, Li, Xia, Xu, Wu, Chang, Sun, Li, and Sui}]{dong-etal-2024-survey}
Qingxiu Dong, Lei Li, Damai Dai, Ce~Zheng, Jingyuan Ma, Rui Li, Heming Xia, Jingjing Xu, Zhiyong Wu, Baobao Chang, Xu~Sun, Lei Li, and Zhifang Sui. 2024.
\newblock \href {https://doi.org/10.18653/v1/2024.emnlp-main.64} {A survey on in-context learning}.
\newblock In \emph{Proceedings of the 2024 Conference on Empirical Methods in Natural Language Processing}, pages 1107--1128, Miami, Florida, USA. Association for Computational Linguistics.

\bibitem[{Gao et~al.(2023)Gao, Madaan, Zhou, Alon, Liu, Yang, Callan, and Neubig}]{gao2023pal}
Luyu Gao, Aman Madaan, Shuyan Zhou, Uri Alon, Pengfei Liu, Yiming Yang, Jamie Callan, and Graham Neubig. 2023.
\newblock \href {https://arxiv.org/abs/2211.10435} {Pal: Program-aided language models}.
\newblock \emph{Preprint}, arXiv:2211.10435.

\bibitem[{Geva et~al.(2021)Geva, Khashabi, Segal, Khot, Roth, and Berant}]{geva-etal-2021-aristotle}
Mor Geva, Daniel Khashabi, Elad Segal, Tushar Khot, Dan Roth, and Jonathan Berant. 2021.
\newblock \href {https://doi.org/10.1162/tacl_a_00370} {Did aristotle use a laptop? a question answering benchmark with implicit reasoning strategies}.
\newblock \emph{Transactions of the Association for Computational Linguistics}, 9:346--361.

\bibitem[{Gronchi and Giovannelli(2018)}]{gronchi2018dual}
G~Gronchi and F~Giovannelli. 2018.
\newblock \href {https://doi.org/10.3389/fpsyg.2018.01237} {Dual process theory of thought and default mode network: A possible neural foundation of fast thinking}.
\newblock \emph{Frontiers in Psychology}, 9:1237.

\bibitem[{Gu et~al.(2025)Gu, Jiang, Shi, Tan, Zhai, Xu, Li, Shen, Ma, Liu, Wang, Zhang, Wang, Gao, Ni, and Guo}]{gu2025surveyllmasajudge}
Jiawei Gu, Xuhui Jiang, Zhichao Shi, Hexiang Tan, Xuehao Zhai, Chengjin Xu, Wei Li, Yinghan Shen, Shengjie Ma, Honghao Liu, Saizhuo Wang, Kun Zhang, Yuanzhuo Wang, Wen Gao, Lionel Ni, and Jian Guo. 2025.
\newblock \href {https://arxiv.org/abs/2411.15594} {A survey on llm-as-a-judge}.
\newblock \emph{Preprint}, arXiv:2411.15594.

\bibitem[{Han et~al.(2025)Han, Wang, Fang, Zhao, Ma, and Chen}]{han2025token}
Tingxu Han, Zhenting Wang, Chunrong Fang, Shiyu Zhao, Shiqing Ma, and Zhenyu Chen. 2025.
\newblock Token-budget-aware llm reasoning.
\newblock \emph{arXiv preprint arXiv:2412.18547}.

\bibitem[{Hendrycks et~al.(2021)Hendrycks, Burns, Basart, Zou, Mazeika, Song, and Steinhardt}]{hendrycks2021measuring}
Dan Hendrycks, Collin Burns, Steven Basart, Andy Zou, Mantas Mazeika, Dawn Song, and Jacob Steinhardt. 2021.
\newblock \href {https://arxiv.org/abs/2009.03300} {Measuring massive multitask language understanding}.
\newblock \emph{Preprint}, arXiv:2009.03300.

\bibitem[{Huang et~al.(2024)Huang, Zhang, Luck, Bu, Qing, and Cui}]{huang2024agentcodermultiagentbasedcodegeneration}
Dong Huang, Jie~M. Zhang, Michael Luck, Qingwen Bu, Yuhao Qing, and Heming Cui. 2024.
\newblock \href {https://arxiv.org/abs/2312.13010} {Agentcoder: Multi-agent-based code generation with iterative testing and optimisation}.
\newblock \emph{Preprint}, arXiv:2312.13010.

\bibitem[{Kahneman(2011)}]{kahneman2011thinking}
Daniel Kahneman. 2011.
\newblock \emph{Thinking, Fast and Slow}.
\newblock Farrar, Straus and Giroux, New York.

\bibitem[{Li et~al.(2023)Li, Hammoud, Itani, Khizbullin, and Ghanem}]{li2023camel}
Guohao Li, Hasan Abed Al~Kader Hammoud, Hani Itani, Dmitrii Khizbullin, and Bernard Ghanem. 2023.
\newblock \href {https://openreview.net/forum?id=3IyL2XWDkG} {{CAMEL}: Communicative agents for ''mind'' exploration of large language model society}.
\newblock In \emph{Thirty-seventh Conference on Neural Information Processing Systems}.

\bibitem[{Li et~al.(2024)Li, Yuan, Feng, Pan, Wang, Sun, Wang, and Li}]{li2024escape}
Yiwei Li, Peiwen Yuan, Shaoxiong Feng, Boyuan Pan, Xinglin Wang, Bin Sun, Heda Wang, and Kan Li. 2024.
\newblock \href {https://openreview.net/forum?id=ndR8Ytrzhh} {Escape sky-high cost: Early-stopping self-consistency for multi-step reasoning}.
\newblock In \emph{The Twelfth International Conference on Learning Representations}.

\bibitem[{Liu et~al.(2024{\natexlab{a}})Liu, Lin, Hewitt, Paranjape, Bevilacqua, Petroni, and Liang}]{liu-etal-2024-lost}
Nelson~F. Liu, Kevin Lin, John Hewitt, Ashwin Paranjape, Michele Bevilacqua, Fabio Petroni, and Percy Liang. 2024{\natexlab{a}}.
\newblock \href {https://doi.org/10.1162/tacl_a_00638} {Lost in the middle: How language models use long contexts}.
\newblock \emph{Transactions of the Association for Computational Linguistics}, 12:157--173.

\bibitem[{Liu et~al.(2024{\natexlab{b}})Liu, Peng, Du, Yin, Liu, and Zhang}]{liu-etal-2024-era}
Yanming Liu, Xinyue Peng, Tianyu Du, Jianwei Yin, Weihao Liu, and Xuhong Zhang. 2024{\natexlab{b}}.
\newblock \href {https://doi.org/10.18653/v1/2024.acl-long.476} {{ERA}-{C}o{T}: Improving chain-of-thought through entity relationship analysis}.
\newblock In \emph{Proceedings of the 62nd Annual Meeting of the Association for Computational Linguistics (Volume 1: Long Papers)}, pages 8780--8794, Bangkok, Thailand. Association for Computational Linguistics.

\bibitem[{Madaan et~al.(2023)Madaan, Tandon, Gupta, Hallinan, Gao, Wiegreffe, Alon, Dziri, Prabhumoye, Yang, Gupta, Majumder, Hermann, Welleck, Yazdanbakhsh, and Clark}]{madaan2023selfrefine}
Aman Madaan, Niket Tandon, Prakhar Gupta, Skyler Hallinan, Luyu Gao, Sarah Wiegreffe, Uri Alon, Nouha Dziri, Shrimai Prabhumoye, Yiming Yang, Shashank Gupta, Bodhisattwa~Prasad Majumder, Katherine Hermann, Sean Welleck, Amir Yazdanbakhsh, and Peter Clark. 2023.
\newblock \href {https://openreview.net/forum?id=S37hOerQLB} {Self-refine: Iterative refinement with self-feedback}.
\newblock In \emph{Thirty-seventh Conference on Neural Information Processing Systems}.

\bibitem[{Miao et~al.(2024)Miao, Teh, and Rainforth}]{miao2024selfcheck}
Ning Miao, Yee~Whye Teh, and Tom Rainforth. 2024.
\newblock \href {https://openreview.net/forum?id=pTHfApDakA} {Selfcheck: Using {LLM}s to zero-shot check their own step-by-step reasoning}.
\newblock In \emph{The Twelfth International Conference on Learning Representations}.

\bibitem[{OpenAI(2022)}]{openai2022goodhart}
OpenAI. 2022.
\newblock Measuring goodhart's law.
\newblock \url{https://openai.com/index/measuring-goodharts-law/}.

\bibitem[{OpenAI(2024)}]{openai2024gpt4technicalreport}
OpenAI. 2024.
\newblock \href {https://arxiv.org/abs/2303.08774} {Gpt-4 technical report}.
\newblock \emph{Preprint}, arXiv:2303.08774.

\bibitem[{Parashar et~al.(2025)Parashar, Olson, Khurana, Li, Ling, Caverlee, and Ji}]{parashar2025inference}
Shubham Parashar, Blake Olson, Sambhav Khurana, Eric Li, Hongyi Ling, James Caverlee, and Shuiwang Ji. 2025.
\newblock Inference-time computations for llm reasoning and planning: A benchmark and insights.
\newblock \emph{arXiv preprint arXiv:2502.12521}.

\bibitem[{Shinn et~al.(2023)Shinn, Cassano, Gopinath, Narasimhan, and Yao}]{shinn2023reflexion}
Noah Shinn, Federico Cassano, Ashwin Gopinath, Karthik~R Narasimhan, and Shunyu Yao. 2023.
\newblock \href {https://openreview.net/forum?id=vAElhFcKW6} {Reflexion: language agents with verbal reinforcement learning}.
\newblock In \emph{Thirty-seventh Conference on Neural Information Processing Systems}.

\bibitem[{Snell et~al.(2024{\natexlab{a}})Snell, Lee, Xu, and Kumar}]{snell2024scalingllmtesttimecompute}
Charlie Snell, Jaehoon Lee, Kelvin Xu, and Aviral Kumar. 2024{\natexlab{a}}.
\newblock \href {https://arxiv.org/abs/2408.03314} {Scaling llm test-time compute optimally can be more effective than scaling model parameters}.
\newblock \emph{Preprint}, arXiv:2408.03314.

\bibitem[{Snell et~al.(2024{\natexlab{b}})Snell, Lee, Xu, and Kumar}]{snell2024scaling}
Charlie Snell, Jaehoon Lee, Kelvin Xu, and Aviral Kumar. 2024{\natexlab{b}}.
\newblock Scaling llm test-time compute optimally can be more effective than scaling model parameters.
\newblock \emph{arXiv preprint arXiv:2408.03314}.

\bibitem[{Srivastava et~al.(2023)}]{srivastava2023beyondimitationgame}
Aarohi Srivastava et~al. 2023.
\newblock \href {https://arxiv.org/abs/2206.04615} {Beyond the imitation game: Quantifying and extrapolating the capabilities of language models}.
\newblock \emph{Preprint}, arXiv:2206.04615.
\newblock Preprint, arXiv:2206.04615.

\bibitem[{Talmor et~al.(2019)Talmor, Herzig, Lourie, and Berant}]{talmor-etal-2019-commonsenseqa}
Alon Talmor, Jonathan Herzig, Nicholas Lourie, and Jonathan Berant. 2019.
\newblock \href {https://doi.org/10.18653/v1/N19-1421} {{C}ommonsense{QA}: A question answering challenge targeting commonsense knowledge}.
\newblock In \emph{Proceedings of the 2019 Conference of the North {A}merican Chapter of the Association for Computational Linguistics: Human Language Technologies, Volume 1 (Long and Short Papers)}, pages 4149--4158, Minneapolis, Minnesota. Association for Computational Linguistics.

\bibitem[{Touvron et~al.(2023)Touvron, Martin, Stone, Albert, Almahairi, Babaei, Bashlykov, Batra, Bhargava, Bhosale, Bikel, Blecher, Ferrer, Chen, Cucurull, Esiobu, Fernandes, Fu, Fu, Fuller, Gao, Goswami, Goyal, Hartshorn, Hosseini, Hou, Inan, Kardas, Kerkez, Khabsa, Kloumann, Korenev, Koura, Lachaux, Lavril, Lee, Liskovich, Lu, Mao, Martinet, Mihaylov, Mishra, Molybog, Nie, Poulton, Reizenstein, Rungta, Saladi, Schelten, Silva, Smith, Subramanian, Tan, Tang, Taylor, Williams, Kuan, Xu, Yan, Zarov, Zhang, Fan, Kambadur, Narang, Rodriguez, Stojnic, Edunov, and Scialom}]{touvron2023llama}
Hugo Touvron, Louis Martin, Kevin Stone, Peter Albert, Amjad Almahairi, Yasmine Babaei, Nikolay Bashlykov, Soumya Batra, Prajjwal Bhargava, Shruti Bhosale, Dan Bikel, Lukas Blecher, Cristian~Canton Ferrer, Moya Chen, Guillem Cucurull, David Esiobu, Jude Fernandes, Jeremy Fu, Wenyin Fu, Brian Fuller, Cynthia Gao, Vedanuj Goswami, Naman Goyal, Anthony Hartshorn, Saghar Hosseini, Rui Hou, Hakan Inan, Marcin Kardas, Viktor Kerkez, Madian Khabsa, Isabel Kloumann, Artem Korenev, Punit~Singh Koura, Marie-Anne Lachaux, Thibaut Lavril, Jenya Lee, Diana Liskovich, Yinghai Lu, Yuning Mao, Xavier Martinet, Todor Mihaylov, Pushkar Mishra, Igor Molybog, Yixin Nie, Andrew Poulton, Jeremy Reizenstein, Rashi Rungta, Kalyan Saladi, Alan Schelten, Ruan Silva, Eric~Michael Smith, Ranjan Subramanian, Xiaoqing~Ellen Tan, Binh Tang, Ross Taylor, Adina Williams, Jian~Xiang Kuan, Puxin Xu, Zheng Yan, Iliyan Zarov, Yuchen Zhang, Angela Fan, Melanie Kambadur, Sharan Narang, Aurelien Rodriguez, Robert Stojnic, Sergey Edunov, and Thomas
  Scialom. 2023.
\newblock \href {https://arxiv.org/abs/2307.09288} {Llama 2: Open foundation and fine-tuned chat models}.
\newblock \emph{Preprint}, arXiv:2307.09288.

\bibitem[{Trung et~al.(2024)Trung, Zhang, Jie, Sun, Jin, and Li}]{trung-etal-2024-reft}
Luong Trung, Xinbo Zhang, Zhanming Jie, Peng Sun, Xiaoran Jin, and Hang Li. 2024.
\newblock \href {https://doi.org/10.18653/v1/2024.acl-long.410} {{R}e{FT}: Reasoning with reinforced fine-tuning}.
\newblock In \emph{Proceedings of the 62nd Annual Meeting of the Association for Computational Linguistics (Volume 1: Long Papers)}, pages 7601--7614, Bangkok, Thailand. Association for Computational Linguistics.

\bibitem[{Umerenkov et~al.(2023)Umerenkov, Zubkova, and Nesterov}]{umerenkov2023decipheringdiagnoseslargelanguage}
D.~Umerenkov, G.~Zubkova, and A.~Nesterov. 2023.
\newblock \href {https://arxiv.org/abs/2310.01708} {Deciphering diagnoses: How large language models explanations influence clinical decision making}.
\newblock \emph{Preprint}, arXiv:2310.01708.

\bibitem[{Wan et~al.(2024)Wan, Wu, Chen, and Li}]{wan2024dynamic}
Guangya Wan, Yuqi Wu, Jie Chen, and Sheng Li. 2024.
\newblock Dynamic self-consistency: Leveraging reasoning paths for efficient llm sampling.
\newblock \emph{arXiv preprint arXiv:2408.17017}.

\bibitem[{Wang et~al.(2024{\natexlab{a}})Wang, Ma, Feng, Zhang, Yang, Zhang, Chen, Tang, Chen, Lin, Zhao, Wei, and Wen}]{Wang_2024}
Lei Wang, Chen Ma, Xueyang Feng, Zeyu Zhang, Hao Yang, Jingsen Zhang, Zhiyuan Chen, Jiakai Tang, Xu~Chen, Yankai Lin, Wayne~Xin Zhao, Zhewei Wei, and Jirong Wen. 2024{\natexlab{a}}.
\newblock \href {https://doi.org/10.1007/s11704-024-40231-1} {A survey on large language model based autonomous agents}.
\newblock \emph{Frontiers of Computer Science}, 18(6).

\bibitem[{Wang et~al.(2024{\natexlab{b}})Wang, Xu, Zhou, Xiong, and Joty}]{wang2024directjudgementpreferenceoptimization}
Peifeng Wang, Austin Xu, Yilun Zhou, Caiming Xiong, and Shafiq Joty. 2024{\natexlab{b}}.
\newblock \href {https://arxiv.org/abs/2409.14664} {Direct judgement preference optimization}.
\newblock \emph{Preprint}, arXiv:2409.14664.

\bibitem[{Wang et~al.(2025{\natexlab{a}})Wang, Wang, Xue, Pang, Liu, Chen, Qiu, Wong, Ji, and Wong}]{wang2025harnessingreasoningeconomysurvey}
Rui Wang, Hongru Wang, Boyang Xue, Jianhui Pang, Shudong Liu, Yi~Chen, Jiahao Qiu, Derek~Fai Wong, Heng Ji, and Kam-Fai Wong. 2025{\natexlab{a}}.
\newblock \href {https://arxiv.org/abs/2503.24377} {Harnessing the reasoning economy: A survey of efficient reasoning for large language models}.
\newblock \emph{Preprint}, arXiv:2503.24377.

\bibitem[{Wang et~al.(2023)Wang, Wei, Schuurmans, Le, Chi, Narang, Chowdhery, and Zhou}]{wang2023selfconsistency}
Xuezhi Wang, Jason Wei, Dale Schuurmans, Quoc Le, Ed~Chi, Sharan Narang, Aakanksha Chowdhery, and Denny Zhou. 2023.
\newblock \href {https://arxiv.org/abs/2203.11171} {Self-consistency improves chain of thought reasoning in language models}.
\newblock \emph{Preprint}, arXiv:2203.11171.

\bibitem[{Wang et~al.(2024{\natexlab{c}})Wang, Ma, Zhang, Ni, Chandra, Guo, Ren, Arulraj, He, Jiang et~al.}]{wang2024mmlu}
Yubo Wang, Xueguang Ma, Ge~Zhang, Yuansheng Ni, Abhranil Chandra, Shiguang Guo, Weiming Ren, Aaran Arulraj, Xuan He, Ziyan Jiang, et~al. 2024{\natexlab{c}}.
\newblock Mmlu-pro: A more robust and challenging multi-task language understanding benchmark.
\newblock \emph{arXiv preprint arXiv:2406.01574}.

\bibitem[{Wang et~al.(2025{\natexlab{b}})Wang, Liu, Xu, Liang, Chen, He, Song, Yu, Li, Zhang et~al.}]{wang2025thoughts}
Yue Wang, Qiuzhi Liu, Jiahao Xu, Tian Liang, Xingyu Chen, Zhiwei He, Linfeng Song, Dian Yu, Juntao Li, Zhuosheng Zhang, et~al. 2025{\natexlab{b}}.
\newblock Thoughts are all over the place: On the underthinking of o1-like llms.
\newblock \emph{arXiv preprint arXiv:2501.18585}.

\bibitem[{Wei et~al.(2022{\natexlab{a}})Wei, Tay, Bommasani, Raffel, Zoph, Borgeaud, Yogatama, Bosma, Zhou, Metzler, Chi, Hashimoto, Vinyals, Liang, Dean, and Fedus}]{wei2022emergent}
Jason Wei, Yi~Tay, Rishi Bommasani, Colin Raffel, Barret Zoph, Sebastian Borgeaud, Dani Yogatama, Maarten Bosma, Denny Zhou, Donald Metzler, Ed~H. Chi, Tatsunori Hashimoto, Oriol Vinyals, Percy Liang, Jeff Dean, and William Fedus. 2022{\natexlab{a}}.
\newblock \href {https://openreview.net/forum?id=yzkSU5zdwD} {Emergent abilities of large language models}.
\newblock \emph{Transactions on Machine Learning Research}.
\newblock Survey Certification.

\bibitem[{Wei et~al.(2022{\natexlab{b}})Wei, Wang, Schuurmans, Bosma, Xia, Chi, Le, Zhou et~al.}]{wei2022chain}
Jason Wei, Xuezhi Wang, Dale Schuurmans, Maarten Bosma, Fei Xia, Ed~Chi, Quoc~V Le, Denny Zhou, et~al. 2022{\natexlab{b}}.
\newblock Chain-of-thought prompting elicits reasoning in large language models.
\newblock \emph{Advances in neural information processing systems}, 35:24824--24837.

\bibitem[{Wu et~al.(2025{\natexlab{a}})Wu, Wan, Li, Zhao, Ma, Ye, Pop, Zhang, and Chen}]{wu2025wisemindrecontextualizingaiknowledgeguided}
Yuqi Wu, Guangya Wan, Jingjing Li, Shengming Zhao, Lingfeng Ma, Tianyi Ye, Ion Pop, Yanbo Zhang, and Jie Chen. 2025{\natexlab{a}}.
\newblock \href {https://arxiv.org/abs/2502.20689} {Wisemind: Recontextualizing ai with a knowledge-guided, theory-informed multi-agent framework for instrumental and humanistic benefits}.
\newblock \emph{Preprint}, arXiv:2502.20689.

\bibitem[{Wu et~al.(2025{\natexlab{b}})Wu, Wang, Du, Jegelka, and Wang}]{wu2025when}
Yuyang Wu, Yifei Wang, Tianqi Du, Stefanie Jegelka, and Yisen Wang. 2025{\natexlab{b}}.
\newblock When more is less: Understanding chain-of-thought length in llms.
\newblock \emph{arXiv preprint arXiv:2502.07266}.

\bibitem[{Yang et~al.(2025)Yang, Lee, Nowak, and Papailiopoulos}]{yang2025towards}
Wenkai Yang, Kangwook Lee, Robert~D Nowak, and Dimitris Papailiopoulos. 2025.
\newblock Towards thinking-optimal scaling of test-time compute for llm reasoning.
\newblock \emph{arXiv preprint arXiv:2502.18080}.

\bibitem[{Yao et~al.(2024{\natexlab{a}})Yao, Yu, Zhao, Shafran, Griffiths, Cao, and Narasimhan}]{yao2024tree}
Shunyu Yao, Dian Yu, Jeffrey Zhao, Izhak Shafran, Tom Griffiths, Yuan Cao, and Karthik Narasimhan. 2024{\natexlab{a}}.
\newblock Tree of thoughts: Deliberate problem solving with large language models.
\newblock \emph{Advances in Neural Information Processing Systems}, 36.

\bibitem[{Yao et~al.(2023)Yao, Zhao, Yu, Du, Shafran, Narasimhan, and Cao}]{yao2023react}
Shunyu Yao, Jeffrey Zhao, Dian Yu, Nan Du, Izhak Shafran, Karthik Narasimhan, and Yuan Cao. 2023.
\newblock \href {https://arxiv.org/abs/2210.03629} {{ReAct}: Synergizing reasoning and acting in language models}.
\newblock In \emph{International Conference on Learning Representations (ICLR)}.

\bibitem[{Yao et~al.(2024{\natexlab{b}})Yao, Li, and Zhao}]{yao-etal-2024-got}
Yao Yao, Zuchao Li, and Hai Zhao. 2024{\natexlab{b}}.
\newblock \href {https://doi.org/10.18653/v1/2024.findings-naacl.183} {{G}o{T}: Effective graph-of-thought reasoning in language models}.
\newblock In \emph{Findings of the Association for Computational Linguistics: NAACL 2024}, pages 2901--2921, Mexico City, Mexico. Association for Computational Linguistics.

\bibitem[{Zhang et~al.(2023)Zhang, Li, Cui, Cai, Liu, Fu, Huang, Zhao, Zhang, Chen, Wang, Luu, Bi, Shi, and Shi}]{zhang2023sirens}
Yue Zhang, Yafu Li, Leyang Cui, Deng Cai, Lemao Liu, Tingchen Fu, Xinting Huang, Enbo Zhao, Yu~Zhang, Yulong Chen, Longyue Wang, Anh~Tuan Luu, Wei Bi, Freda Shi, and Shuming Shi. 2023.
\newblock \href {https://arxiv.org/abs/2309.01219} {Siren's song in the ai ocean: A survey on hallucination in large language models}.
\newblock \emph{Preprint}, arXiv:2309.01219.

\bibitem[{Zhou et~al.(2023)Zhou, Sch{\"a}rli, Hou, Wei, Scales, Wang, Schuurmans, Cui, Bousquet, Le, and Chi}]{zhou2023leasttomost}
Denny Zhou, Nathanael Sch{\"a}rli, Le~Hou, Jason Wei, Nathan Scales, Xuezhi Wang, Dale Schuurmans, Claire Cui, Olivier Bousquet, Quoc~V Le, and Ed~H. Chi. 2023.
\newblock \href {https://openreview.net/forum?id=WZH7099tgfM} {Least-to-most prompting enables complex reasoning in large language models}.
\newblock In \emph{The Eleventh International Conference on Learning Representations}.

\end{thebibliography}

\appendix
\section{Appendix}

\subsection{Package Used and Code}
\label{appendix:code}
For generating LLMs responses and parsing out answers, we utilize packages "langchain" ,"langchain\_openai", "langchain\_anthropic", "langchain\_community", and "langchain\_core" offered by Langchain\footnote{\url{https://www.langchain.com/}}. In addition, we use "pandas" for data processing, "matplotlib" and "seaborn" for visualization, and "numpy" for basic mathematical manipulation. 

Here is the code link for our GitHub:
\url{https://github.com/wan19990901/CoT_rerailer}

\subsection{Experiments Details}
\label{appendix:data}
\subsubsection{Data and Models}
We source our evaluation data from over 20 categories across standard benchmarks:
\begin{itemize}
\item \textbf{Mathematical Reasoning}: MathQA \citep{amini2019mathqa}, GSM8K \citep{cobbe2021training}, MATH \citep{hendrycks2021measuring}
\item \textbf{Symbolic Reasoning}: CoinFlip \citep{wei2022chain}, selected tasks from BigBench \citep{srivastava2023beyondimitationgame}
\item \textbf{Commonsense Reasoning}: CommonsenseQA \citep{talmor-etal-2019-commonsenseqa}, StrategyQA \citep{geva-etal-2021-aristotle}, MMLU \citep{hendrycks2021measuring}, MMLU-Professional \citep{wang2024mmlu}
\end{itemize}
\textbf{Models:} We utilize three state-of-the-art LLMs:
\begin{itemize}
\item Claude-3.5-Sonnet \cite{claude}
\item Llama-3.1 70B \cite{touvron2023llama}
\item GPT-4o-mini \cite{openai2024gpt4technicalreport}
\end{itemize}
For all experiments, we use a temperature of 0.5 to balance exploration and consistency.
\textbf{Baselines:} We compare against several standard prompting methods:
\begin{itemize}
\item Zero-shot and few-shot Chain-of-Thought (CoT)
\item Least-to-Most
\item Self-Consistency (SC) with $n$ = 20 samples
\item Chain-of-Verification (CoVe)
\end{itemize}

\noindent\textbf{Data Sources}
Our experiments draw from multiple standardized benchmarks to ensure comprehensive evaluation across different reasoning types:
\begin{itemize}
    \item \textbf{BigBench}: We use multiple tasks including Date Understanding, Logical Deduction, and Penguins in a Table, which test different aspects of reasoning capabilities; we also take subset of data from \textbf{BIG-Bench Hard (BBH)}: to focus on Boolean Expressions, Object Counting, Date Understanding, Sports Understanding, and Temporal Sequences tasks for evaluation as alternative datasets for symbolic reasoning
    
    \item \textbf{Various Mathematical Reasoning Benchmark}: We incorporate GSM8K, MathQA, and MATH datasets for comprehensive mathematical reasoning assessment.
    
    \item \textbf{Various Benchmark Commonsense Reasoning}: We use CommonsenseQA and StrategyQA for evaluating practical reasoning and strategic thinking capabilities.
    
    \item \textbf{MMLU}: We utilize both general and professional subjects from MMLU, including medicine, law, engineering, and various scientific disciplines.
    
    \item \textbf{Additional Benchmarks}: We also dataset including CoinFlip for symbolic reasoning evaluation.
\end{itemize}

\noindent\textbf{Task Categories}
Our evaluation framework spans three broad categories of reasoning:

\begin{itemize}
    \item \textbf{Mathematical Reasoning} (1500 questions):
    \begin{itemize}
        \item Core Mathematics (MATH, MathQA)
        \item Applied Mathematics (GSM8K)
        \item Professional Mathematics (MMLU-Mathematics)
        \item Statistical Reasoning
        \item Abstract Algebra
        \item Quantitative Problem-Solving
    \end{itemize}
    
    \item \textbf{Symbolic Reasoning} (1000 questions):
    \begin{itemize}
        \item Formal Logic
        \item Boolean Expressions (BBH)
        \item Temporal Reasoning
        \item Programming Concepts
        \item Physical Systems
        \item Circuit Analysis
    \end{itemize}
    
    \item \textbf{Commonsense Reasoning} (1500 questions):
    \begin{itemize}
        \item General Common Sense (CommonsenseQA)
        \item Strategic Thinking (StrategyQA)
        \item Temporal Understanding
        \item Ethical Reasoning
        \item Professional Judgment
        \item Scientific Reasoning
    \end{itemize}
\end{itemize}

\noindent\textbf{Evaluation Protocol}
We conduct experiments using a standardized protocol across all models and tasks:
\begin{itemize}
    \item Each experiment is repeated 10 times to ensure statistical reliability
    \item Standard deviation is calculated across all metrics
    \item Token usage is carefully tracked for efficiency analysis
    \item Performance is measured both within categories and across the entire test set
\end{itemize}

The categorization is designed to reflect the diversity and scope of the subjects our models are evaluated against, ensuring a comprehensive assessment across a wide array of knowledge domains. To obtain LLM RP for the experiment, we spend roughly \$2000 USD and 250 hrs in total for LLM API usage, combining both Claude and OPENAI.

\subsection{Prompts}
\label{appendix:prompts}
In this section, we included our prompts for each component and their corresponding results. 

\subsubsection{Zero-Shot CoT Generator}
\begin{tcolorbox}
\textbf{System Message:} \textit{You are an expert in \{subject\}. Break down this problem step-by-step, showing all calculations and intermediate reasoning. Let's solve this systematically.}

\textbf{Human:} \textit{\{question\}}
\end{tcolorbox}

\subsubsection{Least-to-Most CoT Generator}
\begin{tcolorbox}
\textbf{System Message:} \textit{You are an expert problem solver. Break down complex problems into simpler sub-problems, solving them in order of increasing difficulty.}

\textbf{Human:} \textit{In a school fundraiser, tickets cost \$5 each. If they sold 145 tickets and the school keeps 70\% of the money, how much does the school keep?}

\textbf{Assistant:} Let me break this down:
1. Sub-questions (from simplest to most complex):
   - How much money did they collect in total?
   - What does 70\% mean in decimal form?
   - How much money does the school keep?

2. Let's solve each:
   Q1: What's the total money collected from ticket sales?
   A1: 145 tickets × \$5 = \$725 total

   Q2: What's 70\% as a decimal?
   A2: 70\% = 70/100 = 0.70

   Q3: How much does the school keep?
   A3: \$725 × 0.70 = \$507.50

Therefore, the school keeps \$507.50.
\end{tcolorbox}

\subsubsection{Three-Shot CoT Generator}
\begin{tcolorbox}
\textbf{System Message:} \textit{You are an expert problem solver. For each problem, break down your reasoning step-by-step, showing all calculations and intermediate thoughts.}

\textbf{Human:} \textit{What is 15\% of 80?}

\textbf{Assistant:} Let me solve this step-by-step:
1. Convert 15\% to decimal: 15\% = 0.15
2. Multiply: 80 × 0.15 = 12
Therefore, 15\% of 80 is 12.

\textbf{Human:} \textit{If John has 3 apples and Mary has twice as many, how many apples do they have together?}

\textbf{Assistant:} Let me solve this step-by-step:
1. John's apples = 3
2. Mary's apples = 2 × 3 = 6
3. Total apples = John's + Mary's = 3 + 6 = 9
Therefore, they have 9 apples together.

\textbf{Human:} \textit{If 6 cookies cost \$2, how much do 15 cookies cost?}

\textbf{Assistant:} Let me solve this step-by-step:
1. Find cost per cookie: \$2 ÷ 6 = \$0.33...
2. Multiply by 15: \$0.33... × 15 = \$5
Therefore, 15 cookies cost \$5.

\textbf{Human:} \textit{\{question\}}
\end{tcolorbox}

\subsubsection{Chain-of-Verification}

\begin{tcolorbox}[title={CoVe Framework}]

\textbf{1. Baseline Response}
\begin{itemize}
    \item \textbf{Query:} \{question\}
    \item \textbf{System:} Generate initial response
    \item \textbf{Output:} [Initial model response]
\end{itemize}

\textbf{2. Plan Verifications}
\begin{itemize}
    \item \textbf{Context:} [Initial response]
    \item \textbf{System:} Generate verification questions for each claim
    \item \textbf{Output:} [List of verification questions]
\end{itemize}

\textbf{3. Execute Verifications}
\begin{itemize}
    \item \textbf{System:} Answer each question independently
    \item \textbf{Format:}
    \begin{itemize}
        \item Q1: [Verification question]
        \item A1: [Independent answer]
        \item Qx: ...
        \item Ax: ...
    \end{itemize}
\end{itemize}

\textbf{4. Cross-Check Analysis}
\begin{itemize}
    \item \textbf{Original:} [Claim from initial response]
    \item \textbf{Verification:} [Q\&A result]
    \item \textbf{Status:} [CONSISTENT/INCONSISTENT/PARTIAL]
    \item \textbf{Evidence:} [Explanation]
\end{itemize}

\textbf{5. Final Verified Response}
\begin{itemize}
    \item \textbf{Query:} [Original question]
    \item \textbf{System:} Generate revised response using verified claims
    \item \textbf{Output:} [Final verified response]
\end{itemize}

\end{tcolorbox}

\subsubsection{Rerailer Prompt Implementation}
\begin{tcolorbox}
\textbf{System Message:} \textit{You are an expert problem solver implementing a multi-step reasoning verification system. For each step, generate and evaluate multiple solution paths to ensure stability.}

\paragraph{Stage 1: Multiple Solution Generation}
For each reasoning step $i$:
\begin{itemize}
    \item \textbf{Input:} Question $x$ and previous steps $s_{1:i-1}$
    \item \textbf{Prompt:} \textit{Generate $k$ diverse solutions for this step. Ensure solutions are meaningfully different.}
    \item \textbf{Output:} $k$ candidate solutions $\{s_i^1, \ldots, s_i^k\}$
\end{itemize}

\paragraph{Stage 2: Pairwise Evaluation}
For each pair of solutions $(s_i^p, s_i^q)$:
\begin{itemize}
    \item \textbf{Prompt:} \textit{Compare these two solutions:}
    \begin{itemize}
        \item Solution 1: [First solution]
        \item Solution 2: [Second solution]
        \item \textit{Which solution is more logical and why?}
    \end{itemize}
    \item \textbf{Output:} Binary comparison score $f(s_i^p, s_i^q)$
\end{itemize}

\paragraph{Stage 3: Score Aggregation}
For each solution $s_i^j$:
\begin{itemize}
    \item \textbf{Input:} All pairwise comparison results
    \item \textbf{Computation:} $\sum_{l=1, l\neq j}^k f(s_i^j, s_i^l)$
    \item \textbf{Output:} Aggregate score for each solution
\end{itemize}

\paragraph{Stage 4: Optimal Path Selection}
\begin{itemize}
    \item \textbf{Input:} Aggregated scores for all solutions
    \item \textbf{Selection:} Select $s_i^*$ with highest score
    \item \textbf{Chain Construction:} Build $S^* = \{s_1^*, \ldots, s_n^*\}$
\end{itemize}

\end{tcolorbox}

\begin{tcolorbox}
\textbf{Example Implementation:}

\textbf{Question:} \textit{If a rectangle has width 4cm and length twice its width, what is its area?}

\textbf{Stage 1 - Generate Solutions:}
\begin{itemize}
    \item $s_1^1$: "Width = 4cm, length = 2 × width"
    \item $s_1^2$: "Width = 4cm, length = 8cm (double the width)"
    \item $s_1^3$: "Given w = 4cm, l = 2w = 8cm"
\end{itemize}

\textbf{Stage 2 - Compare:}
\begin{itemize}
    \item $s_1^2$ vs $s_1^1$: $s_1^2$ better (more explicit)
    \item $s_1^3$ vs $s_1^1$: $s_1^3$ better (more systematic)
    \item $s_1^3$ vs $s_1^2$: $s_1^3$ better (shows work)
\end{itemize}

\textbf{Stage 3 - Scores:}
\begin{itemize}
    \item $s_1^1$: 0 wins
    \item $s_1^2$: 1 win
    \item $s_1^3$: 2 wins
\end{itemize}

\textbf{Stage 4 - Select:} $s_1^3$ chosen

\textbf{Final Answer:} Area = length × width = 8cm × 4cm = 32cm²
\end{tcolorbox}

\subsection{Additional Preliminary Findings}
Besides the sampling size findings, the preliminary study also reveals interesting implications for model size and task type as shown in Fig.~\ref{fig:pre_additional}
\label{APP:preliminary}
\subsubsection{Model Size and Consistency}
Our results indicate that the ratio of consistently correct or consistently incorrect cases to inconsistent cases differs across models. Larger models (e.g., gpt-4o) tend to have higher confidence and produce more consistent answers, whether correct or incorrect. Smaller models, such as a llama3-7b variant, exhibit greater variability in answers. Interestingly, while large models may have a high proportion of \emph{Solvable} outcomes, smaller models show a broader spread across consistent and inconsistent categories.

These findings support the hypothesis that consistency level is correlated with model size and training sophistication. Larger models are generally more knowledgeable and stable, reducing the necessity for complex prompting in easily solvable questions but also being less susceptible to improvement on questions they \emph{consistently fail} without external knowledge augmentation.

\subsubsection{Task Type: Commonsense vs. Mathematical Reasoning} When examining commonsense (CM) reasoning and mathematical reasoning tasks, we observe distinct patterns. For CM tasks, the LLM responses are often either consistently correct or consistently incorrect, suggesting that the model either “knows” the fact or does not. The variability in these responses is low, indicating minimal benefit from advanced prompting techniques: if the model’s knowledge is lacking, complex prompts alone are unlikely to help. In contrast, mathematical reasoning tasks show a more substantial spread, with a notable portion falling into the \emph{inconsistently solvable} category. This suggests that the model’s reasoning processes for mathematics are more delicate and prone to occasional errors or hallucinations, even when partial capability exists. Here, advanced prompting techniques, such as step-by-step reasoning or self-checking mechanisms, can meaningfully increase consistency and correctness. This reinforces the rationale for using specialized derailing strategies—introducing additional reasoning steps, verification routines, or multi-stage solution prompts.
\begin{figure*}[th!]
    \centering    \includegraphics[width=\linewidth,height=10cm,keepaspectratio]{figures/preliminary.jpg}
    \caption{Consensus level and question type across different models and categories.}
    \label{fig:pre_additional}
\end{figure*}

\subsection*{Pass Rate Analysis}
\label{APP:derailer}
\begin{figure*}[th!]
    \centering    \includegraphics[width=\linewidth,height=10cm,keepaspectratio]{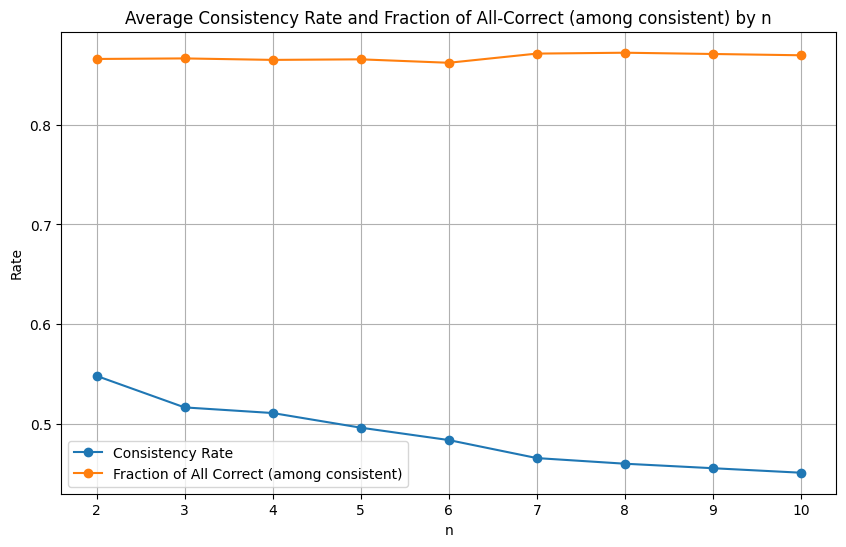}
    \caption{pass rate of all of data of varying N}
    \label{fig:pass_all}
\end{figure*}

\begin{figure*}[th!]
    \centering    \includegraphics[width=\linewidth,height=10cm,keepaspectratio]{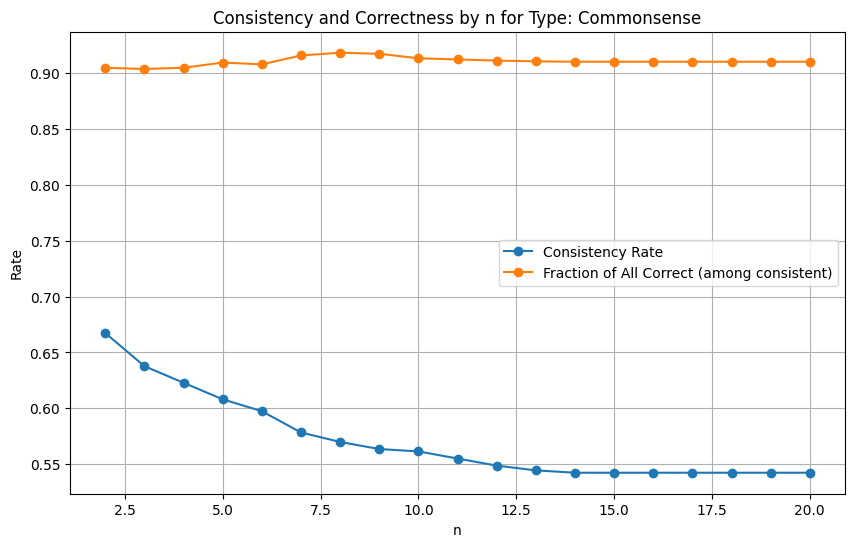}
    \caption{pass rate of Commonsense reasoning of varying N}
    \label{fig:pass_cs}
\end{figure*}

\begin{figure*}[th!]
    \centering    \includegraphics[width=\linewidth,height=10cm,keepaspectratio]{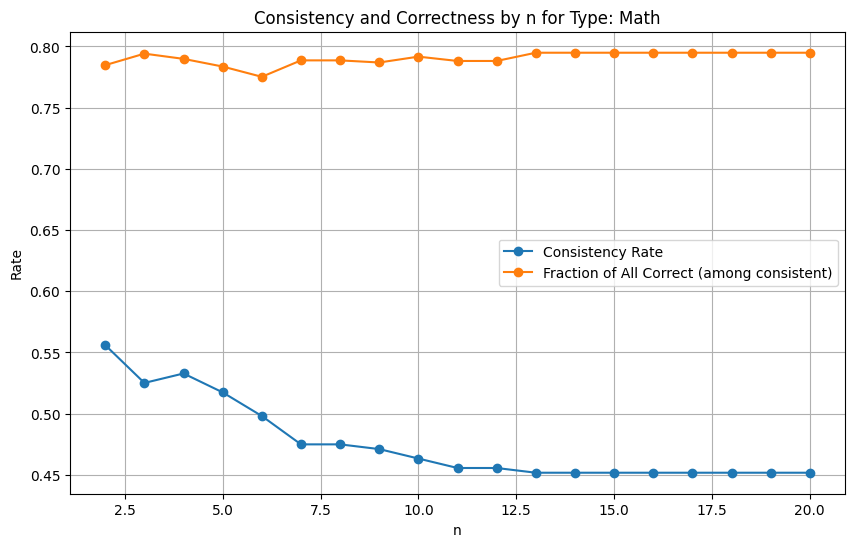}
    \caption{pass rate of all Math data of varying N}
    \label{fig:pass_math}
\end{figure*}

\begin{figure*}[th!]
    \centering    \includegraphics[width=\linewidth,height=10cm,keepaspectratio]{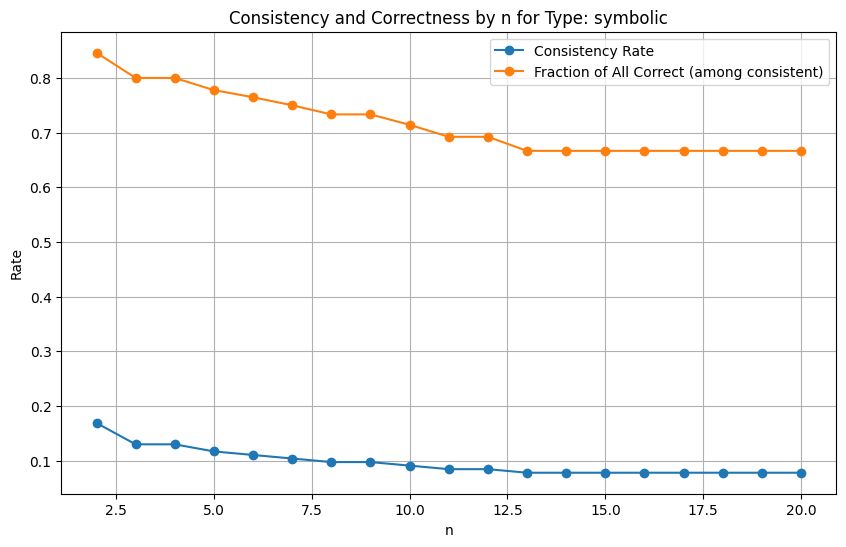}
    \caption{pass rate of Symbolic Reasoning data of varying N}
    \label{fig:pass_sym}
\end{figure*}

We provide a detailed analysis of Derailer's pass rates (proportion of questions flagged for iterative prompting) across different sample sizes and reasoning categories. As shown in Fig.\ref{fig:pass_all}, the overall pass rates remain remarkably stable across different values of $N$. When breaking down by reasoning types, we observe some variation in absolute pass rates between commonsense (Fig.\ref{fig:pass_cs}), mathematical (Fig.\ref{fig:pass_math}), and symbolic reasoning (Fig.\ref{fig:pass_sym}), reflecting the different difficulty levels of these domains. However, the stability pattern holds across all categories - increasing the number of samples beyond 5 does not significantly change which questions are flagged for additional reasoning. This consistency supports our main paper's findings about the efficiency of small sample sizes for detecting reasoning stability.

\label{app:pass_rate}

\subsection{Error Analysis and Case Study}
\label{app:error_analysis}

\begin{figure*}[th!]
    \centering    \includegraphics[width=\linewidth,height=10cm,keepaspectratio]{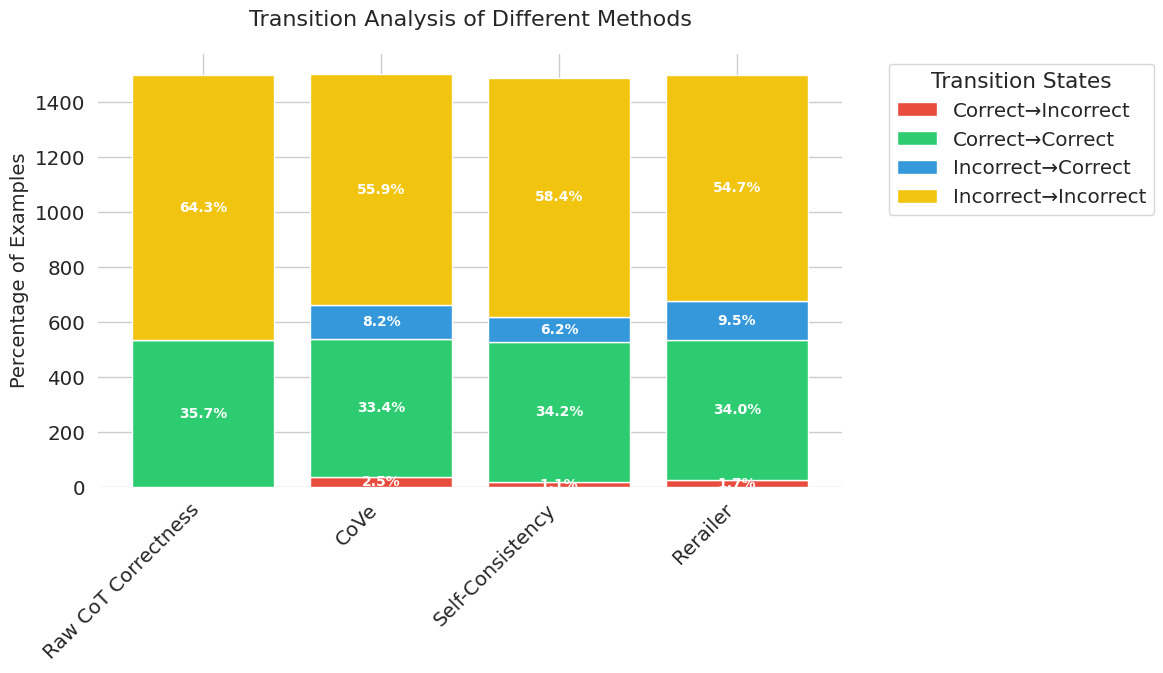}
    \caption{Correction ratio of various iterative prompting method with respect to CoT data after Derailer}
    \label{fig:correction}
\end{figure*}

\subsubsection{Error Correction Analysis}

We provided a detailed analysis on how the instablizd reasoning gets fixed after Derailer. The Figure \ref{fig:correction} demonstrates Rerailer's balanced approach to handling unstable reasoning patterns. While maintaining comparable stability for correct answers (Correct→Correct ~34\%), Rerailer achieves the highest rate of error correction (Incorrect→Correct at 9.5\%) while showing the lowest rate of destabilizing correct answers (Correct→Incorrect at 1.8\%). This indicates our proposed Rerailer's ability to effectively identify and fix genuine errors without over-correcting stable reasoning paths, outperforming both CoVe and Self-Consistency while preserving good efficiency.

\begin{figure*}[h]
\begin{center}
\includegraphics[width=\linewidth]{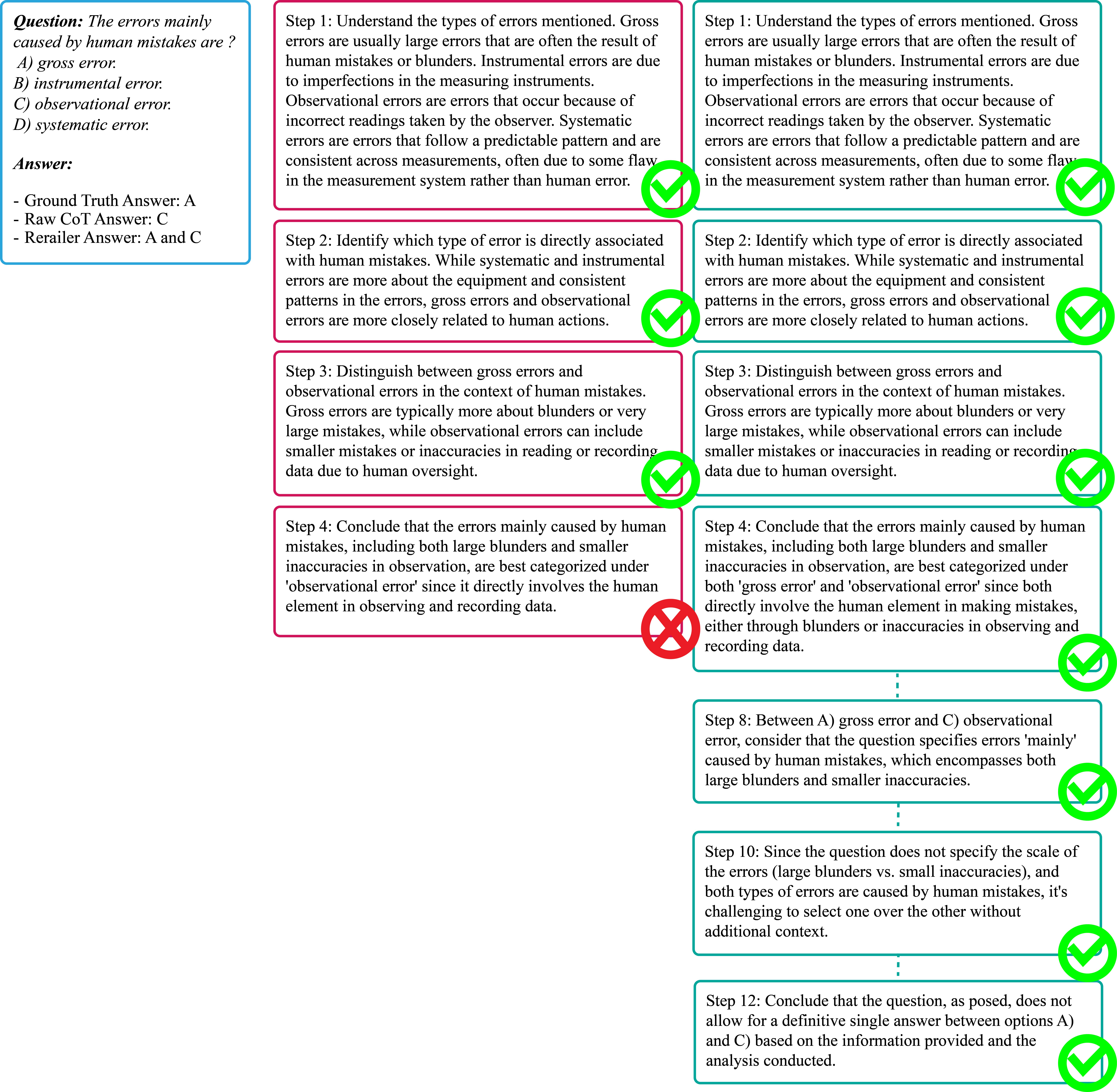}

\end{center}
\caption{Error analysis-Lacking Background Information Global Facts Problem. The questions and answers, which were retrieved from the MMLU dataset, are exhibited in the blue box. The red boxes are steps generated via the baseline CoT method and the green boxes are the corrected CoT from Rerailer.}
\label{lack_of_background_info}
\end{figure*}

In our experiment, mainly three types of typical errors occurred with potentially downgrading our performance including wrong ground-truth, lack of background information, and ambiguous questions. The detailed example were show in Fig.~\ref{lack_of_background_info}

\begin{figure*}[h]
\begin{center}
\includegraphics[width=\linewidth]{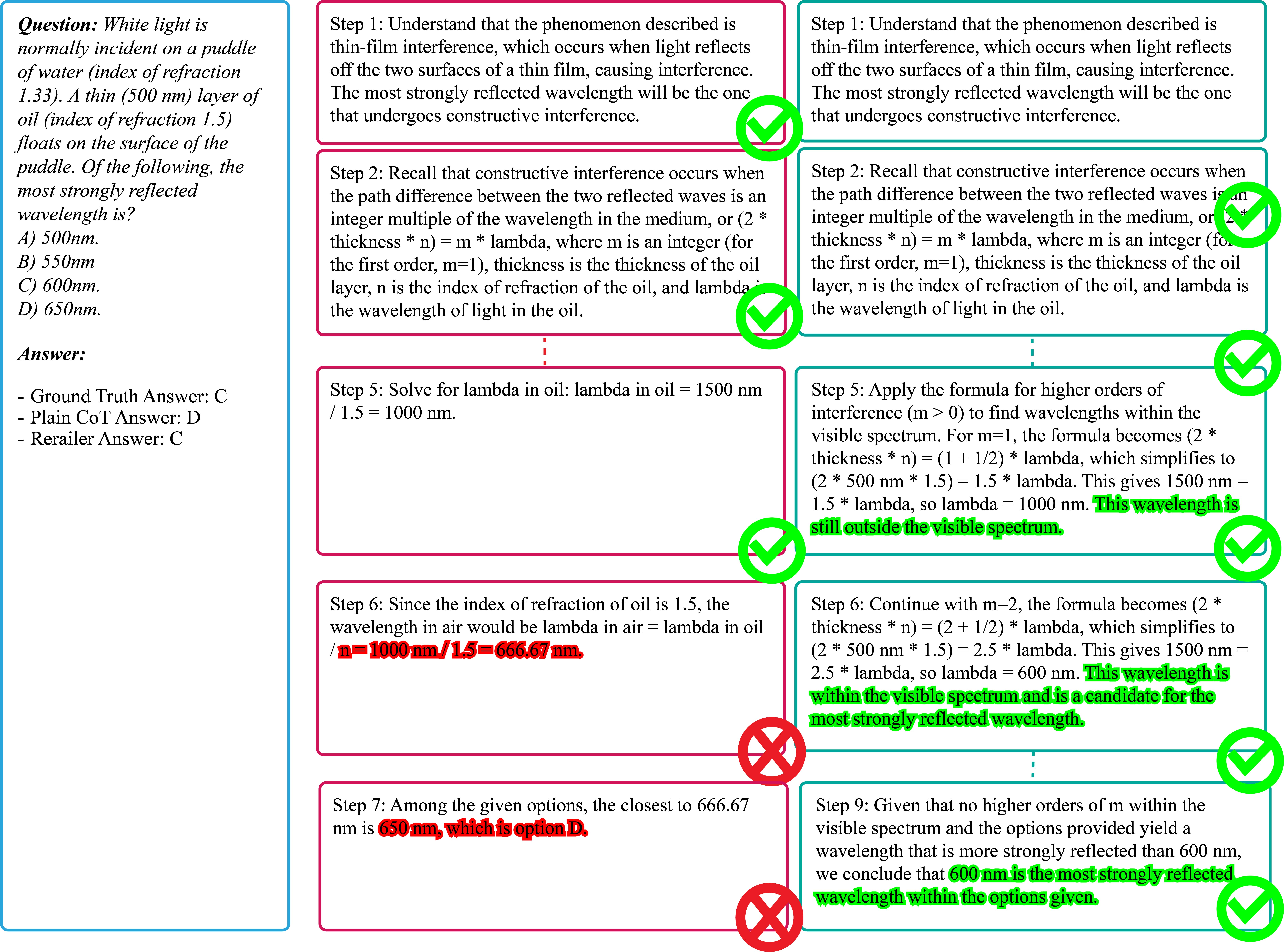}

\end{center}
\caption{Derailer-Rerailer Solving a Physics Question. The questions and answers, which were retrieved from the MMLU dataset, are exhibited in the blue box. The red boxes are steps generated via the baseline CoT method and the green boxes are the corrected RP from Rerailer. Mistakes are highlighted in red and corrections are highlighted in green.}
\label{postive_sample_physics}
\end{figure*}

\begin{figure*}[h]
\begin{center}
\includegraphics[width=\linewidth]{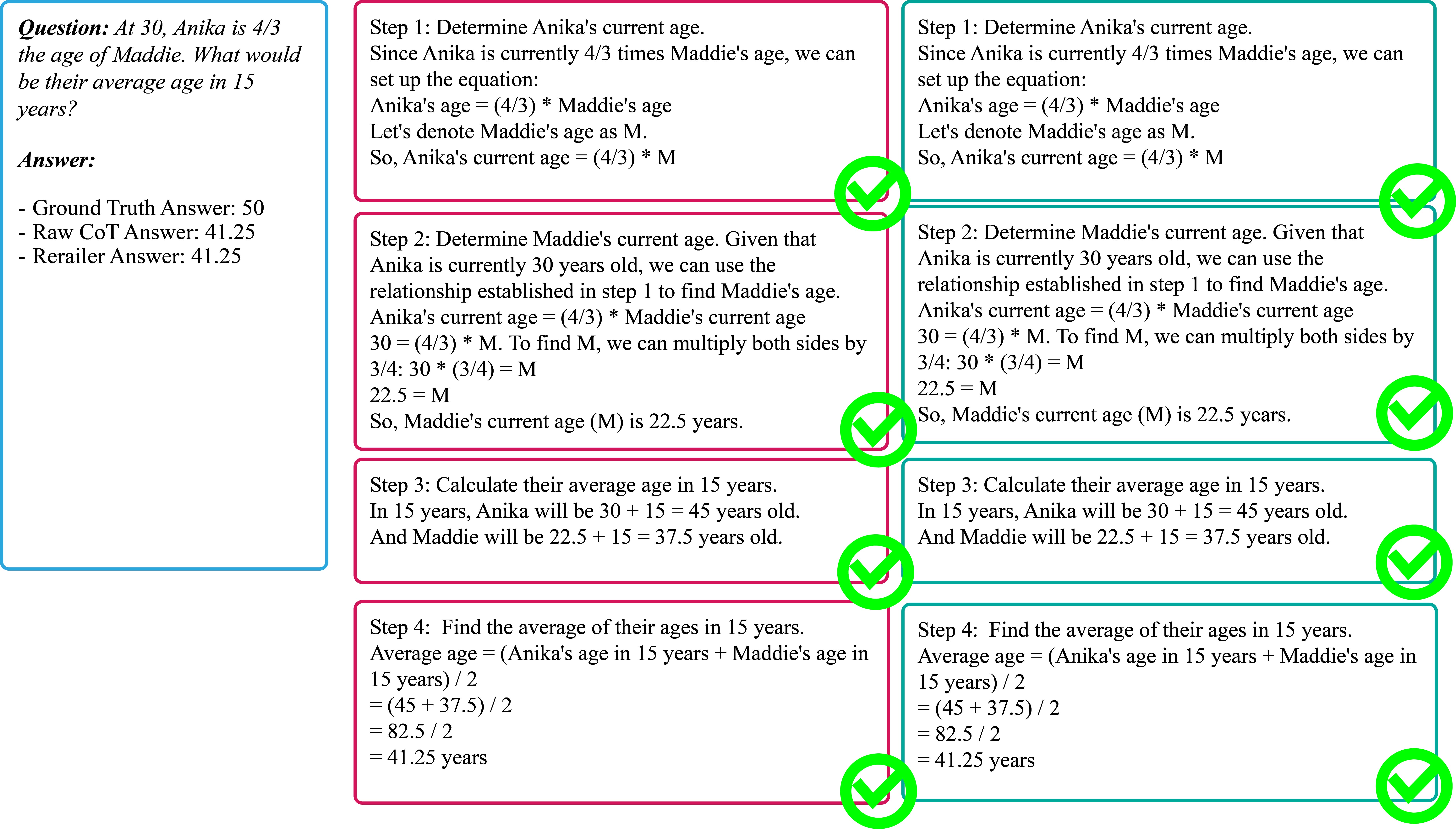}

\end{center}
\caption{Error analysis-Wrong Ground Truth Math Problem. The questions and answers, which were retrieved from the GSM8K dataset, are exhibited in the blue box. The red boxes are steps generated via the baseline CoT method and the green boxes are the corrected RP from Rerailer.}
\label{wrong_gt}
\end{figure*}

\end{document}